\documentclass[lettersize,journal]{IEEEtran}
\usepackage{amsmath,amsfonts}
\usepackage{algorithmic}
\usepackage{algorithm}
\usepackage{array}
\usepackage[caption=false,font=normalsize,labelfont=sf,textfont=sf]{subfig}
\usepackage{textcomp}
\usepackage{stfloats}
\usepackage{url}
\usepackage{verbatim}
\usepackage{graphicx}
\usepackage{cite}
\usepackage{booktabs}
\usepackage{color}

\newcommand{\resp}[1]{\textcolor{black}{#1}}

\newcommand{\minor}[1]{\textcolor{black}{#1}}

\begin{document}

\title{Tuning-Free Adaptive Style Incorporation for Structure-Consistent Text-Driven Style Transfer}
\author{Yanqi Ge$^*$, Jiaqi Liu$^*$, Qingnan Fan$^\dag$, Xi Jiang, Ye Huang, Shuai Qin, Hong Gu, Wen Li, and~Lixin Duan$^\dag$ 

\IEEEcompsocitemizethanks{
\IEEEcompsocthanksitem This work is supported by National Natural Science Foundation of China (No. 82121003), and was completed during Yanqi Ge's internship at VIVO. 
\IEEEcompsocthanksitem Yanqi Ge, Wen Li and Lixin Duan are with the School of Computer Science and Engineering, University of Electronic Science and Technology of China (UESTC), Chengdu 611731, China. Lixin Duan is also with the Sichuan Provincial Key Laboratory for Human Disease Gene Study and the Center for Medical Genetics, Department of Laboratory Medicine, Sichuan Academy of Medical Sciences and Sichuan Provincial People’s Hospital, UESTC, Chengdu 610032, China. E-mail: geyanqiqi@gmail.com; wenli.vision@gmail.com; lxduan@uestc.edu.cn.
\IEEEcompsocthanksitem Ye Huang is with Shenzhen Institute for Advanced Study, University of Electronic Science and Technology of China, Shenzhen 518109, China. E-mail: edward.ye.huang@qq.com.
\IEEEcompsocthanksitem Jiaqi Liu and Xi Jiang are with the Department of Computer Science and Engineering, Southern University of Science and Technology, Shenzhen 518055, China. E-mail:  liujq32021@mail.sustech.edu.cn; jiangx2020@mail.sustech.edu.cn.
\IEEEcompsocthanksitem Qingnan Fan, Shuai Qin and Hong Gu are with VIVO, Hangzhou 310023, China. E-mail: fqnchina@gmail.com; shuai.qin@vivo.com; guhong@vivo.com.
\IEEEcompsocthanksitem $^*$ Equal contribution. $\dag$ Corresponding authors.
}
}

\markboth{Journal of \LaTeX\ Class Files,~Vol.~14, No.~8, August~2021}%
{Shell \MakeLowercase{\textit{et al.}}: A Sample Article Using IEEEtran.cls for IEEE Journals}


\maketitle

\begin{abstract}
Text-driven style transfer methods leveraging diffusion models have shown impressive creativity, yet they still face challenges in maintaining consistent structure and content preservation.
Existing methods often directly concatenate the content and style prompts for a prompt-level style injection.
However, this coarse-grained style injection strategy inevitably leads to structural deviations in the stylized images.
This poses a significant obstacle for professional artists and creators seeking precise artistic editing.
In this work, we strive to attain a harmonious balance between content preservation and style transformation.
We propose Adaptive Style Incorporation (ASI), to achieve fine-grained feature-level style incorporation. It consists of the Siamese Cross-Attention~(SiCA) to decouple the single-track cross-attention to a dual-track structure to obtain separate content and style features, and the Adaptive Content-Style Blending (AdaBlending) module to couple the content and style information from a structure-consistent manner. Experimentally, our method exhibits much better performance in both structure preservation and stylized effects. 
\end{abstract}


\begin{figure*}[t]
    \centering
    \includegraphics[width=1\linewidth]{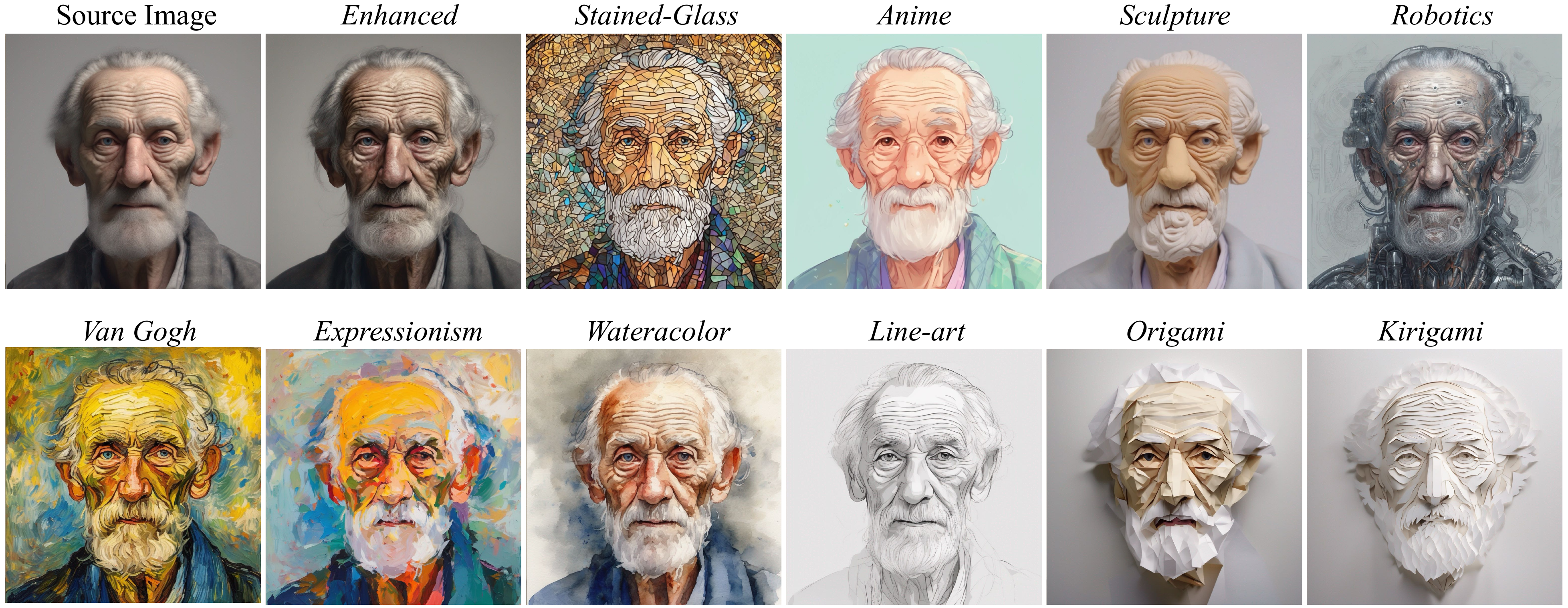} 
    \caption{We propose Adaptive Style Incorporation (ASI), a tuning-free diffusion-based style transfer method that enables versatile text-guided stylization for the source image. Our stylized results exhibit high consistency to the structure and semantics of the source image, while significantly changing their image style following the style prompt.  }
    \label{fig:teaser}
\end{figure*}

\section{Introduction}
\label{sec:intro}
\IEEEPARstart{W}{ith} the increasing prominence of text-to-image (T2I)~\cite{saharia2022photorealistic,ramesh2022hierarchical,rombach2022high} models featuring billion-parameter architectures trained on extensive text-image datasets, the prospect of translating our imaginative ideas into high-quality images through language becomes readily apparent.
One thrilling application is the image style transfer via textual inputs.
However, due to the text-image misalignment problem in the T2I models, even minor modifications to the textual prompt may result in unforeseen structure distortion in the output image.
This poses a significant obstacle for professional creators who aim for artistic renderings while precisely preserving the semantics, structure, and context of the source image.

In this work, we strive to attain a harmonious balance between content preservation and style transformation.
To achieve this objective, numerous efforts have been made to incorporate additional conditional controls, such as structural guidance signals~\cite{ye2023ip,zhang2023adding} and editing masks~\cite{gafni2022make,avrahami2023blended}.
To be specific, BlendDiffusion~\cite{avrahami2022blended, avrahami2023blended} and SpaText~\cite{avrahami2023spatext} train diffusion models that support a user-provided mask as input to achieve regional generation and editing.
ControlNet~\cite{zhang2023adding} and Make-A-Scene~\cite{gafni2022make} finetune diffusion models conditioned on the segmentation masks to ensure invariance of semantics in the editing images. 
However, these methods require extensive training on large-scale text-image pairs and fall short of achieving fine-grained control over the structural layout, thereby hindering intuitive and easily deployable text-driven style transfer.

Recently, several works~\cite{hertz2022prompt, tumanyan2023plug,cao2023masactrl, alaluf2023cross, geyer2023tokenflow} demonstrate that the attention features in diffusion models implicitly encode fine-grained structure information, and enables text-driven style transfer without the requirement for model tuning. 
This greatly reduces the cost of professional creation using the diffusion model. 
Among all these works, Prompt-to-Prompt (P2P)~\cite{hertz2022prompt} and Plug-and-Play (PnP)~\cite{tumanyan2023plug} share the most similarity to ours methodologically.
P2P proposes operating the cross-attention map from the source image diffusion process, allowing various editing capabilities while preserving the source image content.
PnP proposes manipulating the self-attention map from the source image diffusion process, enabling better retention of fine-grained structural details.

However, the aforementioned approaches choose to directly concatenate the content prompt (representing the source image) and style prompt, to achieve prompt-level style injection. Nevertheless, due to the well-known text-image misalignment issue~\cite{huang2023t2i} in T2I models, the style prompt often contains non-style information, leading to unavoidable structure distortions and semantic drift in the stylized image for P2P and PnP. 

Unlike previous studies that commonly inject styles on the coarse-grained prompt level, we propose Adaptive Style Incorporation (ASI) to incorporate the style information on the fine-grained feature level.
In ASI, we first present the Siamese Cross-Attention (SiCA) that decouples the common single-track cross-attention into a dual-track architecture, which utilizes the shared \textit{Query} features to simultaneously query two sets of \textit{Key} and \textit{Value} features from the content prompt and style prompt individually, as depicted in Fig.~\ref{fig:pipeline} (b).
Subsequently, we introduce a tuning-free Adaptive Content-Style Blending (AdaBlending) module to parse style features that do not disrupt the structure of content features and incorporate them into the content features.
Thanks to the fine-grained style incorporation, the ASI mechanism can preserve better structure consistency between content and stylized images.

Our contributions can be summarized as follows. 
\begin{itemize}
\item We propose a novel solution to the text-driven style transfer task, namely Adaptive Style Incorporation (ASI), to achieve feature-level fine-grained style incorporation. It preserves better image structure while enabling effective style transfer.
\item We propose the Siamese Cross-Attention (SiCA) and Adaptive Content-Style Blending (AdaBlending) modules to form the complete ASI mechanism. It requires no training or fine-tuning but rather leverages a pre-trained text-to-image diffusion model.
\item Experimental results show that our proposed method effectively embeds style attributes while preserving the structural layout of the content image, regardless of whether it is applied to real or generated images.

\end{itemize}

\section{Related Work}

\subsection{Diffusion Models}
The Diffusion Model~\cite{sohl2015deep} was initially employed to mitigate Gaussian noise applied continuously to training images and can be conceived as a series of denoising autoencoders. The advent of Denoising Diffusion Probabilistic Models~\cite{ho2020denoising} propelled the utilization of diffusion models for image generation into the mainstream. Subsequent advancements, exemplified by Denoising Diffusion Implicit Models~\cite{song2020denoising}, substantially accelerated sampling speeds while effecting directed generation from random noise to samples, facilitating subsequent research in image editing. Dhariwal and Nichol first introduced Classifier Guidance Diffusion~\cite{dhariwal2021diffusion}, substantiating the superiority of diffusion models over GANs in image generation tasks. Classifier-Free Diffusion Guidance~\cite{ho2021classifier} then achieved a balance between sample quality and diversity during image generation. Building upon prior work, GLIDE~\cite{nichol2022glide} initiated using text as a conditional guide for image generation in a classifier-free manner. DALLE-2~\cite{ramesh2022hierarchical}, building on the foundation laid by GLIDE, incorporated CLIP as a guiding factor, bringing diffusion models into the public eye. Subsequently, the field of image generation based on diffusion models experienced a surge, with projects such as Imagen~\cite{saharia2022photorealistic} and Stable Diffusion~\cite{rombach2022high} emerging, prompting the gradual application of diffusion models in a broader range of tasks, including object detection~\cite{chen2023diffusiondet}, style transfer~\cite{wang2023stylediffusion}, image editing~\cite{li2024blip,brooks2023instructpix2pix,cheng2023general} and semantic segmentation~\cite{zbinden2023stochastic}. 
\resp{With the release of ControlNet~\cite{zhang2023adding} and IP-Adapter~\cite{ye2023ip}, the use of more diverse and richer conditions for image editing and customized image generation is gaining widespread exploration and application.}

\subsection{Style Transfer}
Style transfer has garnered significant attention from researchers alongside the development of generative models. Early style transfer algorithms~\cite{gatys2016neural}, such as AdaIN~\cite{huang2017arbitrary}, primarily relied on optimization-based style transfer methods applied to the VGG architecture~\cite{simonyan2014very}. With the advent of Generative Adversarial Networks (GANs)~\cite{goodfellow2014generative}, GANs have been employed in training style transfer models tailored to specific domains, such as ink paintings~\cite{he2018chipgan}, landscape art~\cite{xue2021end}, cartoons~\cite{wang2020learning}, and portraiture~\cite{yi2020unpaired}. In addition, representative works such as CycleGAN~\cite{zhu2017unpaired}, StarGAN~\cite{choi2018stargan}, and StyleGAN~\cite{karras2019style} are devoted to the research of multi-domain style transfer. Currently, within the realm of style transfer research built upon the burgeoning diffusion model, there is a predominant focus on multi-domain style transfer studies. Research involving style transfer using diffusion models is akin to studies based on GANs, relying predominantly on loss functions for training. Representative works in this domain include StyleDiffusion~\cite{wang2023stylediffusion}, UCAST~\cite{zhang2023unified}, CycleDiffusion~\cite{wu2023latent}, and InST~\cite{zhang2023inversion}, among others. A limited amount of research has focused on tuning-free style transfer. Some studies, such as StyleAligned~\cite{hertz2023style}, investigate style generation using images as style references. Plug-and-Play~\cite{tumanyan2023plug}, on the other hand, explores style generation using text as a reference, and this study is the closest to our paper. 
However, these methods ignore injecting styles in a structurally consistent manner, thus disturbing the structural consistency between the stylized image and the source image.
Compared to prompt-level coarse-grained one-shot style injection, we propose ASI which enables feature-level fine-grained style incorporation in a structure-consistent manner, achieving a good balance between stylization and structural preservation compared to other SOTA methods.

\section{Preliminaries}
The diffusion model~\cite{sohl2015deep} is a recently prominent probabilistic generative model, which accomplishes the image generation task by progressively predicting and removing Gaussian noise from a noisy image. The mathematical significance of the diffusion model is to simulate a distribution $p_\theta(x_0)$ approximate to the data distribution $q(x_0)$. Typically, diffusion models consist of two main processes: the forward process and the reverse denoising process. The forward process of noise injection is as follows,

\begin{gather}
    q(x_t|x_{t-1})=\mathcal{N}(\sqrt{1-\beta_t}x_{t-1},\beta_t\mathbf{I}), \\
    x_t = \sqrt{1-\beta_t}x_{t-1}+\sqrt{\beta_t}\epsilon, \epsilon \sim \mathcal{N}(0,\mathit{I}),
\end{gather}
where $t\in [0, T]$, $T$ is a preset maximum number of sampling steps, while the values of $\alpha_t$ are in a fixed variance schedule, $\alpha_0=1$ and $\lim_{t \to \infty}\alpha_t=0$. In practice, the forward noise injection process is a Markovian process, which means that given $x_0$, it is possible to obtain the noisy result for any number of steps directly,

\begin{gather}
    q(x_t|x_0)=\mathcal{N}(\sqrt{\bar{\alpha}_t}x_0;(1-\bar{\alpha}_t)\mathbf{I}), \\
    x_t = \sqrt{\alpha_t}x_0+\sqrt{1-\alpha_t}\epsilon, \epsilon \sim \mathcal{N}(0,\mathit{I}),
\end{gather}
where $\alpha_t = 1-\beta_t$ and $\bar{\alpha}_t=\prod_{s=1}^t \alpha_s$.

The reverse denoising process begins with $x_T$ as the starting point and involves sampling from the posteriors $q(x_{t-1}|x_t)$ to perform continuous noise prediction. DDPM~\cite{ho2020denoising} sets the noise variance to a fixed value of $\sigma^2\mathbf{I}$, allowing the Gaussian transformation $p_\theta(x_{t-1}|x_t)$ to be learned through a trainable $\mu_\theta(x_t,t)$. Hence, the reverse denoising process can be represented as follows,

\begin{gather}
    p_\theta(x_{t-1}|x_t)=\mathcal{N}(\mu_\theta(x_t,t),\sigma^2_t\mathbf{I}), \\
    x_{t-1}=\frac{1}{\sqrt{\alpha_t}}(x_t-\frac{1-\alpha_t}{\sqrt{1-\bar{\alpha}_t}}\mu_\theta(x_t,t))+\sigma_t\mathbf{z},
\end{gather}
where $z$ is Gaussian noise independent of $x_t$, and $\mu_\theta(x_t,t)$ is a noise approximator $\epsilon_\theta(x_t,t)$ optimized to mimic the forward noise $\epsilon$:
\begin{equation}
    \underset{\theta}{min}||\mu_\theta(x_t,t)-\epsilon||^2.
\end{equation}

DDIM~\cite{song2020denoising} introduces an alternative non-Markovian process, allowing the denoising process to be different. Given timestep $x_t$ and $\epsilon_\theta(x_t,t)$ at timestep $t$, DDIM can predict $x_0=f_\theta(x_t,t)$ and $x_{t-1}$ as follows,
\begin{gather}
    f_\theta(x_t,t)=\frac{x_t-\sqrt{1-\bar{\alpha}_t}\epsilon_\theta(x_t,t)}{\sqrt{\bar{\alpha}_t}},\\
    x_{t-1}=\sqrt{\bar{\alpha}_{t-1}}f_\theta(x_t,t)+\sqrt{1-\bar{\alpha}_{t-1}-\sigma^2_t\epsilon_\theta(x_t,t)+\sigma_t\mathbf{z}}.\label{eq:ddim}
\end{gather}

One of the distinctive features of DDIM is that changing $\sigma_t$ in Eq.~\ref{eq:ddim} can lead to different inverse processes. When $\sigma_t=0$, it essentially turns the mapping from latent to image into a fixed process, supporting perfect inversion of images. This is the basis for our acquisition of image latent.

\begin{figure*}[t]
    \centering
    \includegraphics[width=1\linewidth]{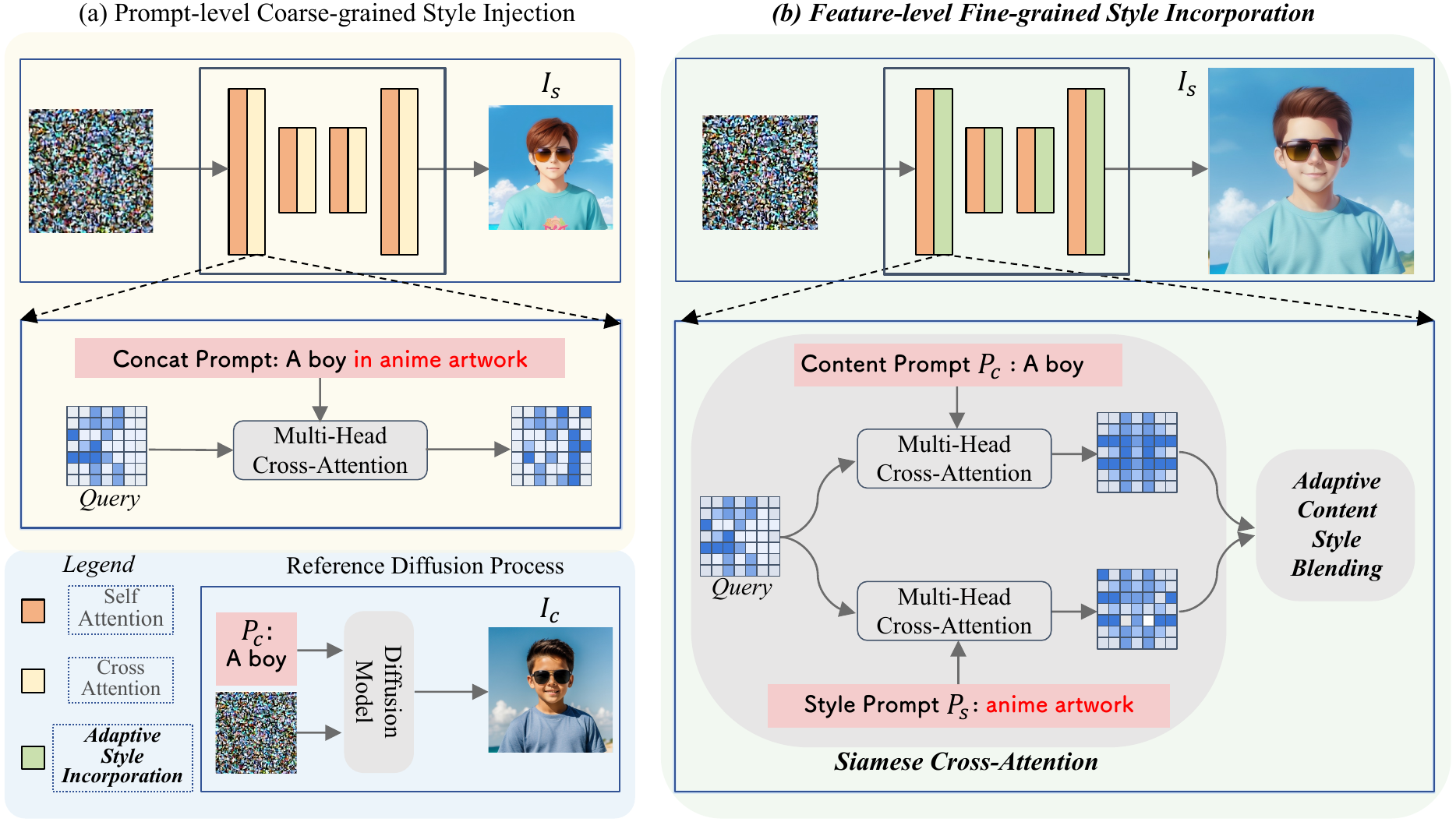}
    
    \caption{ (a) Due to the text-image misalignment in T2I models, directly concatenating the content prompt with the style prompt (i.e., prompt-level coarse-grained style injection) will introduce non-style information into the style transfer process, resulting in unavoidable structural and semantic drift in the stylized image, such as the hair of the boy and the pattern on his clothes.
    (b) We propose Adaptive Style Incorporation (ASI), which consists of siamese cross-attention and adaptive content style blending modules. ASI explicitly parses style information that does not disrupt the structure of content features and incorporates it into the content features to achieve feature-level fine-grained style incorporation.
    }

    \label{fig:pipeline}
\end{figure*}

\begin{figure*}[t]
    \centering
    \hspace{0.5cm}\includegraphics[width=0.91\linewidth]{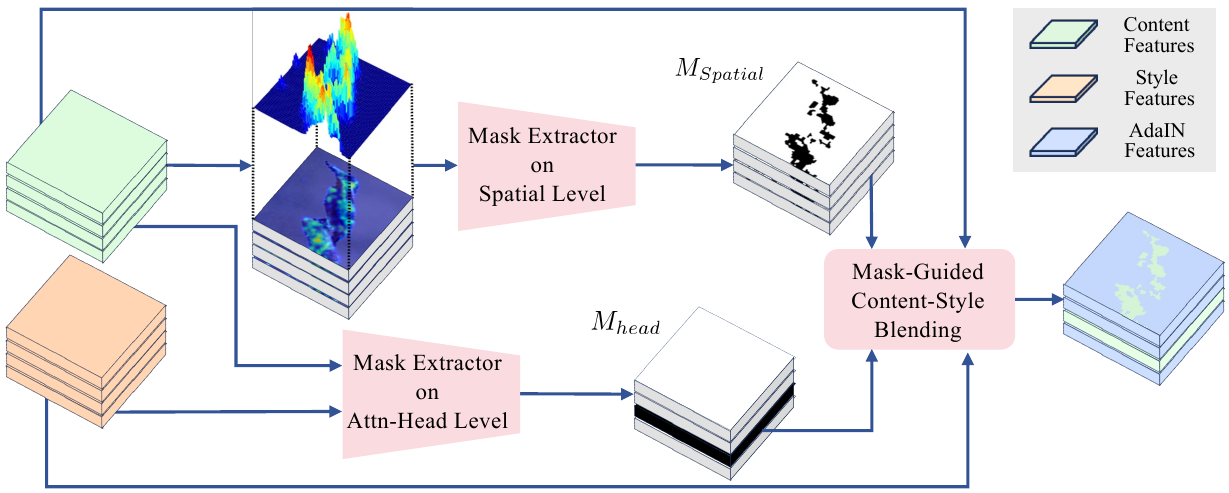}
    \caption{Overall architecture of our proposed Adaptive Content-Style Blending (AdaBlending) module. AdaBlending achieves mask-guided fine-grained style incorporation through the proposed attention-head level and spatial level mask extractors, followed by the content-style blending operation in the structure-consistent manner.}
    \label{fig: adablending}
\end{figure*}

\section{Proposed Method}

Given a latent representation of a real or synthetic source content image $I_c$ along with its corresponding content prompt $P_c$ and a style prompt $P_s$, 
our objective is to generate a target stylized image $I_s$ that maintains structural and semantic consistency with $I_c$ while exhibiting the stylistic characteristics described by $P_s$.

As discussed in Sec.~\ref{sec:intro}, existing methods~\cite{hertz2022prompt, tumanyan2023plug, meng2021sdedit} commonly directly concatenate the content prompt and style prompt (e.g., a boy in anime artwork) and transform to key-value feature pair to achieve prompt-level style injection.
We question this coarse-grained form of prompt-level style injection.
Our motivation stems from the fact that if we do not reuse the spatial features of the reference diffusion process, and simply add a style token to the content prompt, the text-image misalignment problem in the T2I model will often result in unpredictable structural drift in the stylized image $I_s$.
In this sense, we gain greater control over the results if we explicitly parse style information that does not disrupt the structure of content features and incorporate it into content features.

For this purpose, we propose the Adaptive Style Incorporation (ASI) to convert the existing cross-attention module to achieve structure-consistent feature-level fine-grained style transfer.
The overall architecture of the proposed ASI is depicted in Fig.~\ref{fig:pipeline}, which is composed of a Siamese Cross-Attention (SiCA) strategy and Adaptive Content-Style Blending (AdaBlending) module.
We first introduce SiCA which employs a dual-track architecture to obtain independent content and style feature-level information (Sec.~\ref{sec: sica}).
Then, we propose the AdaBlending module to implement mask-guided fine-grained style incorporation from a structure-consistent perspective (Sec.~\ref{sec: adablending}).
The proposed method can be applied to style transfer for both real and generated images without requiring training or fine-tuning, achieving a highly structurally consistent stylized image with the content image.

\subsection{Siamese Cross-Attention}
\label{sec: sica}
As depicted in Fig.~\ref{fig:pipeline} (b), we first present a Siamese Cross-Attention (SiCA) that decouples the common single-track cross-attention into a dual-track architecture, where the two branches of the architecture share the parameters of the replaced cross-attention. We call these two branches the ``style branch'' and ``content branch''. In addition, we formalize the output features $F_s$ and $F_c$ as follows,
\begin{gather}
    F_s(Q,K_s,V_s) = \text{Softmax}(\frac{QK_s^T}{\sqrt{d}})V_s, \\
    F_c(Q,K_c,V_c) = \text{Softmax}(\frac{QK_c^T}{\sqrt{d}})V_c,
\label{eq:attention}
\end{gather} 
where $Q$ is the shared query features projected from the spatial features, $K_s-V_s$ and $K_c-V_c$ are the key-value feature pairs projected from the textual embedding of style prompt and content prompt 
with corresponding projection matrices, and $d$ is the channel dimension of query features. 

This design aims to empower the query features, which signify the image structure, to independently attend to both the style and content prompt embeddings.
Compared to prompt-level coarse-grained one-shot style injection, we can initially extract the style information from $F_s$ with minimal structural impact on $F_c$ and adaptively incorporate it into $F_c$, achieving feature-level fine-grained style incorporation. We will detail it in the next section.

\subsection{Adaptive Content-Style Blending}
\label{sec: adablending}
We propose a tuning-free Adaptive Content-Style Blending (AdaBlending) module to implement mask-guided fine-grained style incorporation from a structure-consistent perspective.
The overall architecture of the AdaBlending is shown in Fig.~\ref{fig: adablending}. 
AdaBlending first utilizes two parallel mask extractors to parse out the style incorporation region, which is composed of an attention-head mask $M_{head}$ and spatial mask $M_{spatial}$.
The attention-head mask $M_{head}$ ensures that the style injection performed on each attention head is effective, while the spatial mask $M_{spatial}$ prevents style injection on the high-frequency edge information of the content image, thereby preserving the integrity of the content object.

With these masks, AdaBlending restricts the content features $F_c$ to incorporate style features $F_s$ from a structurally consistent perspective.
We first perform logical \texttt{OR} operations on two sets of masks $M_{head}$ and $M_{spatial}$ to obtain the final mask $M_{all}$.
Then, we perform the mask-guided blending between the structured style features powered by adaptive instance normalization operation~\cite{huang2017arbitrary}, and content features to obtain the fused fine-grained stylized result as AdaBlending's output $F_{out}$,
\begin{equation}\label{eq:adain}
\begin{aligned}
    F^i_{out} =& (\sigma(F_s^i)(\frac{F_c^i-\mu(F_c^i)}{\sigma(F_c^i)})+\mu(F_s^i)) \odot M_{all}^i \\
               &+ F_c^i \odot(\textbf{1} - M_{all}^i),
\end{aligned}
\end{equation}
where $\odot$ denotes the inner product operation. 
\resp{the $\mu(\cdot)$ and $\sigma(\cdot)$ represent functions for taking the standard deviation and the mean of features, respectively.}
The next two sections will detail how to generate the $M_{head}$ and $M_{spatial}$.

\subsubsection{Mask Extractor on Attention-Head Level}
\label{sec: attn mask}
Many pioneering works ~\cite{huang2017arbitrary, hertz2023style, ulyanov2016instance, mroueh2019wasserstein, zhou2021domain, liu2021adaattn, li2017demystifying, jing2019neural, zhang2022exact, chung2024style} have demonstrated that the spatial distribution of features effectively represents the style of an image.
Therefore, if an attention head does not exhibit significant distribution differences in $F_s$ and $F_c$, we consider incorporating style information into this attention head ineffective and disrupts the stylized image's image structure.

We measure the distribution difference~\cite{sun2016deep} between each pair of attention heads in $F_s$ and $F_c$ by calculating the distance of covariance $\ell^i$ as follows,

\begin{gather}
  Cov_{s}= {\frac{1}{m-1}}({(F_s^i)^{\top} F_s^i - \frac{1}{m}{({\textbf{1}}^{\top}F_s^i})^{\top}{({\textbf{1}}^{\top}F_s^i})}), \\
  Cov_{c}= {\frac{1}{m-1}}({(F_c^i)^{\top} F_c^i - \frac{1}{m}{({\textbf{1}}^{\top}F_c^i})^{\top}{({\textbf{1}}^{\top}F_c^i})}), \\
  \ell^i= \frac{||Cov_s-Cov_c||^2_F}{4d^2},
  \label{eq:cor_l}
\end{gather}
where $Cov_s$ and $Cov_c$ denote the spatial level covariance of style and content features, respectively. ${\|\cdot\|}^2_F$ represents the squared matrix Frobenius norm, $\textbf{1}$ is a vector with all elements equal to 1, $m$, $d$, $i$ are flattened spatial dimension, channel dimension, and attention-head index, respectively.
Then, we can obtain the attention-head mask~$ M_{head} \in \mathbb{R}^{h\times m \times 1}$ as follows,
\begin{equation}
\label{eq: filter}
    M_{head}^i  =\begin{cases} 
        1, \quad &\text{if} ~~~~\mathop{\arg\max}_{i} \ \ ( \ell^i, n) \\
        0, \quad &\text{otherwise}  \\
    \end{cases}\,  
\end{equation}
where $\mathop{\arg\max}_{i}~( \ell^i, n)$ denotes the return of the index of top-$n$ attention heads with the greatest covariance distances. \resp{$n$ is set to 6 for visual enhancement and to 7 for style transfer in the Stable Diffusion V1 architecture.} $h$ denotes the number of heads of multi-head cross-attention.

\begin{figure}[!t]
    \centering
    \includegraphics[width=0.83\linewidth]{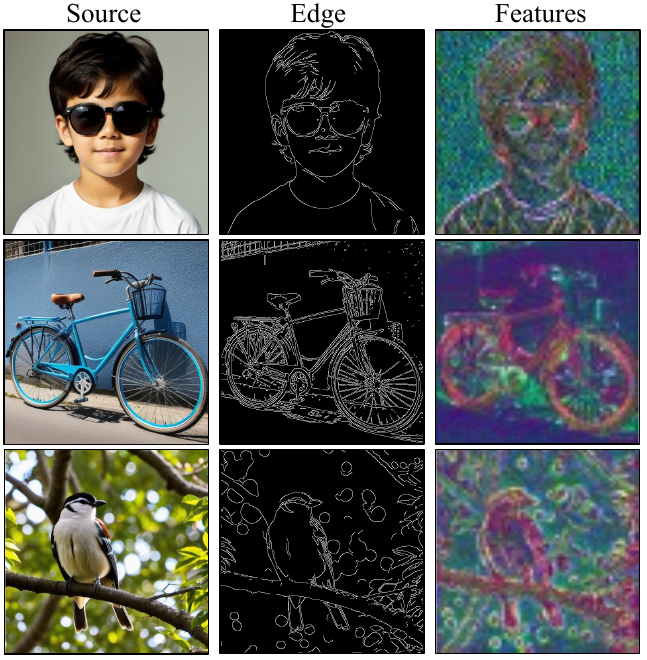}
    \caption{Visualization of the top three leading components of average cross-attention features of the U-Net.}
    \label{fig: motivation2}
\end{figure}

\subsubsection{Mask Extractor on Spatial Level}
\label{sec: spatial mask} 
We further consider how to maintain structure consistency from the spatial level.
Inspired by previous work~\cite{hertz2022prompt,tumanyan2023plug,cao2023masactrl,tang-etal-2023-daam}, the leading principal components of the attention features implicitly encode the image core structure information.
As shown in Fig.~\ref{fig: motivation2}, the highly activated positions basic correspond to the image's high-frequency low-level edge information.

Therefore, we propose a spatial-level mask extractor to achieve style incorporation exclusively in non-structural regions.
The extractor selects criteria points of peak regions using the global max pooling of $F_c$'s first principal component, and the criteria points are denoted as $P \in \mathbb{R}^{h\times m \times 1}$. 
We adopt a non-learnable setting for the extractor and define regions in $F_c$'s first principal component exceeding $\tau = \alpha \cdot P$ as crucial structural regions. $\alpha$ is a constant value and set to $0.7$.

We formulate the spatial mask $M_{spatial} \in \mathbb{R}^{h\times m \times 1}$ at each attention-head as follows,
\begin{equation}
\label{eq: selector}
    M_{spatial}^i  =\begin{cases} 
        1, \quad &\text{if} ~~~~F_c^i < \tau \\
        0, \quad &\text{otherwise}  \\
    \end{cases}\,  
\end{equation}

\begin{figure}[!t]
    \centering
    \includegraphics[width=1\linewidth]{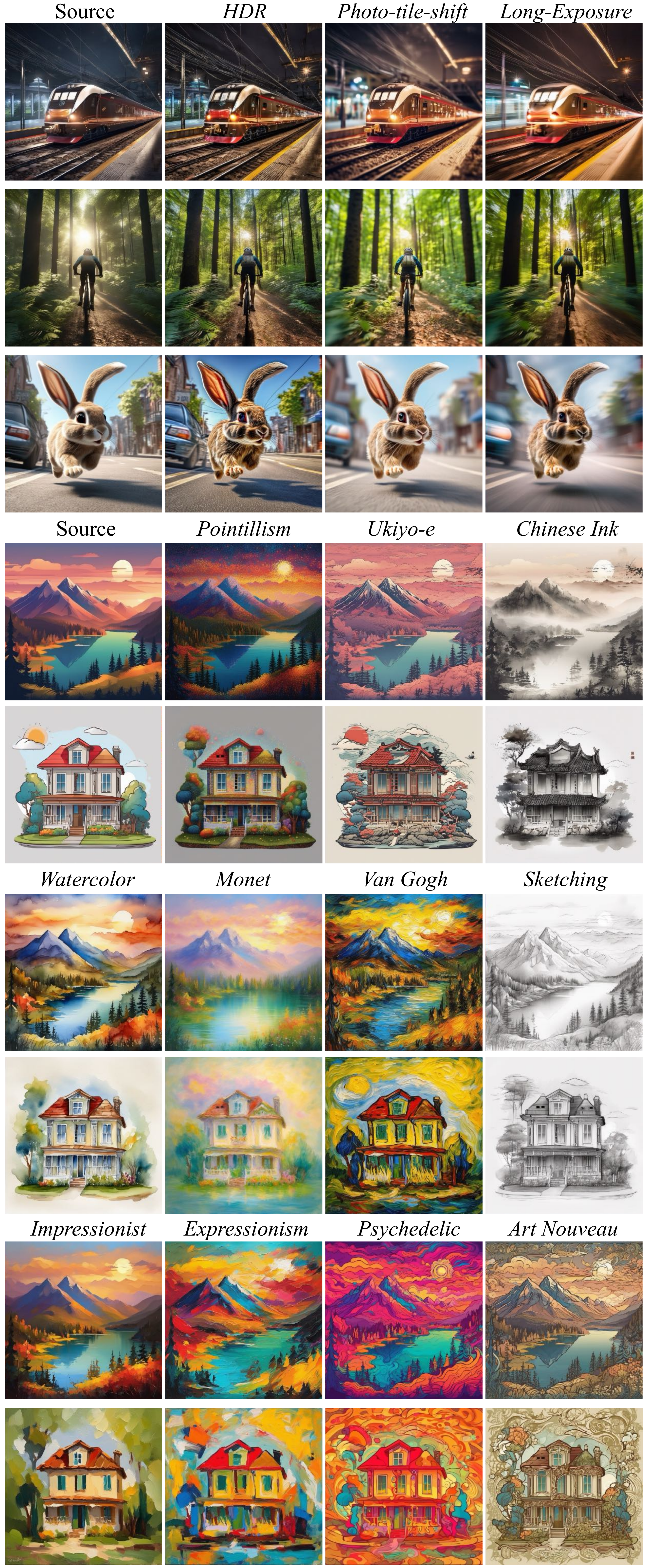}
    \caption{Sample results of our method for transferring both photography and artistic styles. Zoom in for the best view.}
    \label{fig: exp1}
\end{figure}

\section{Experiments}

\minor{In this section, we conduct both qualitative and quantitative evaluations of style transfer and visual enhancement tasks. These evaluations are performed on both real-world images and images generated from given prompts using the Stable Diffusion model.}
We focus our comparisons on SOTA tuning-free text-driven style transfer method based on diffusion models, including FreeDiff~\cite{wu2024freediff}, PnP~\cite{tumanyan2023plug}, P2P~\cite{hertz2022prompt}, and SDEdit~\cite{meng2021sdedit}. 
In addition, we further qualitatively compare our method with other methods that have different experimental settings, including \minor{AdaIN}~\cite{huang2017arbitrary}, AdaAttn~\cite{liu2021adaattn}, Z*~\cite{deng2024z}, DiffStyler~\cite{huang2024diffstyler}, 
\minor{ClipStyler}~\cite{kwon2022clipstyler}, IP-Adapter~\cite{ye2023ip}, and ControlNet~\cite{zhang2023adding}. 
Finally, we ablate each component of our method and show its effectiveness.

Rows 1-3 of Fig. \ref{fig: exp1} present examples of images applied to three classic photographic effects: HDR, photo-tile-shift, and long-exposure. Rows 4-9 of Fig. \ref{fig: exp1} show the results of transfer a single image to various art styles. Our method successfully preserves the structure and natural appearance of the image, while enabling effective style transfer.

\begin{figure*}[t]
    \centering
    \includegraphics[width=1\linewidth]{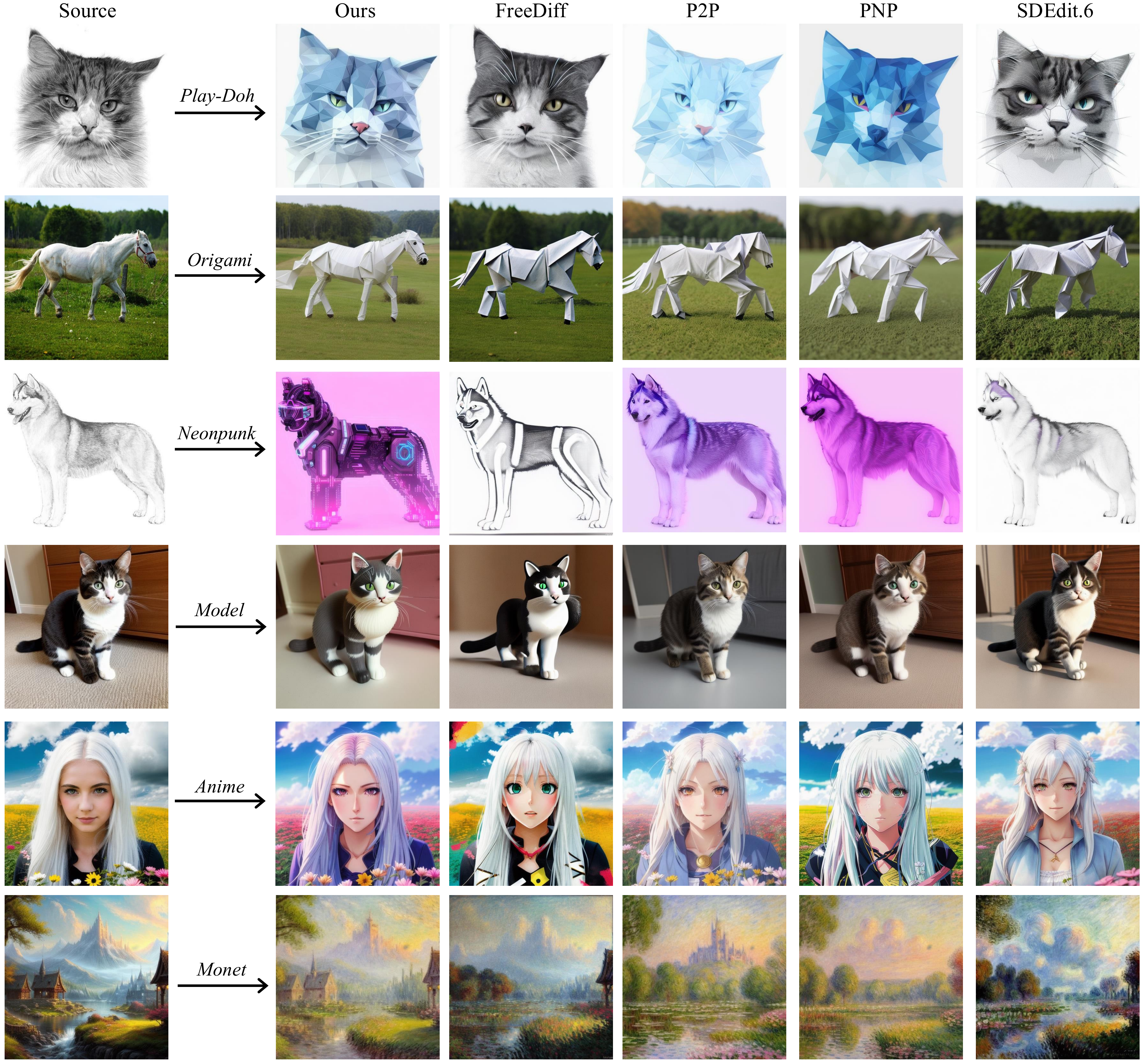}

    \caption{ Comparison of style transfer results using different methods. From left to right: the source content image and text prompt, our results, FreeDiff, P2P, PnP, and SDEdit with $0.6$ denoising strengths. Zoom in for the best view.}
        
    \label{fig:style_transfer}
\end{figure*}

\begin{figure*}[t]
    \centering
    \includegraphics[width=1\linewidth]{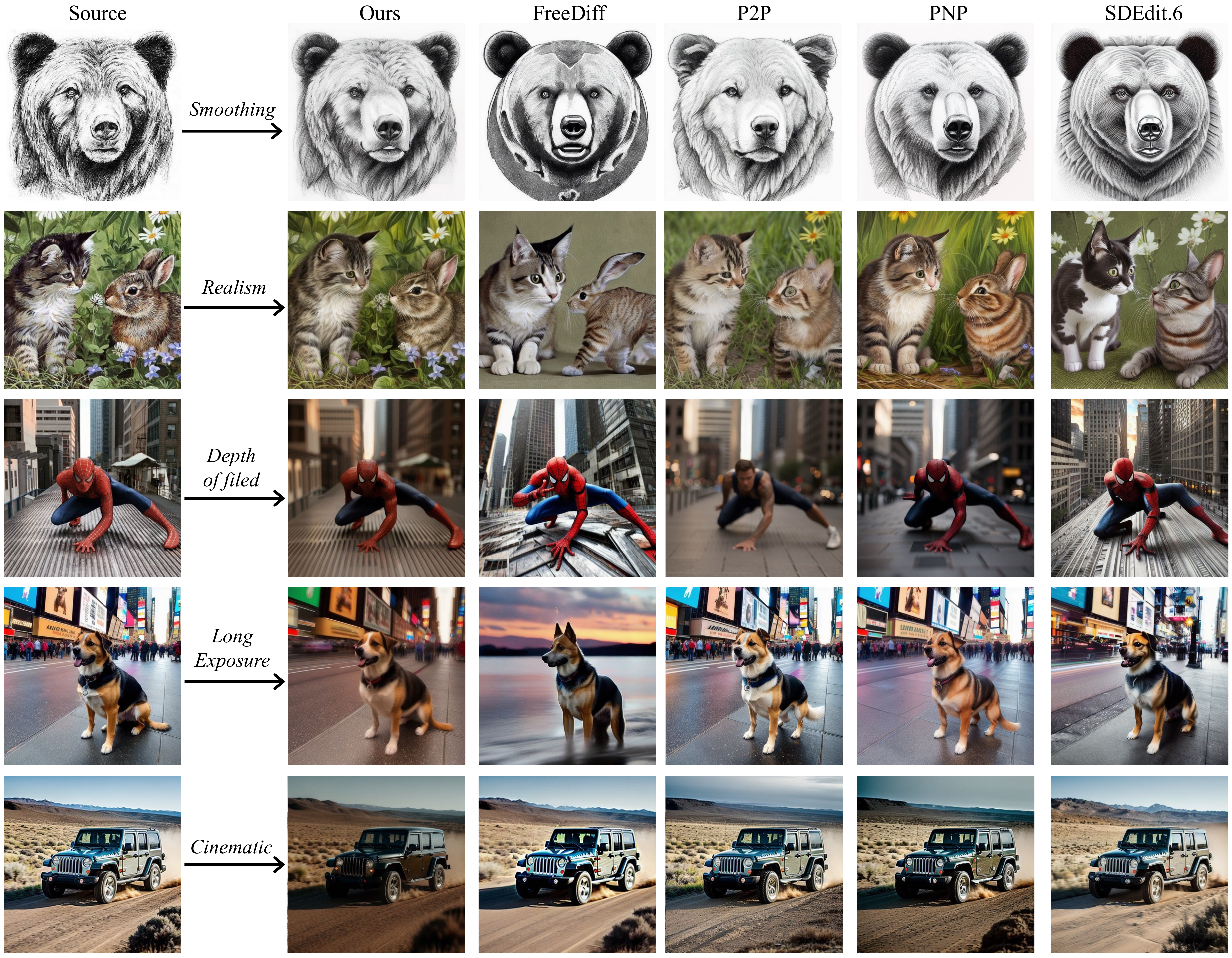}
    \caption{Comparison of visual enhancement results using different methods. From left to right: the source content image, visual enhancement prompts, our results, FreeDiff, P2P, PnP, and SDEdit with $0.6$ denoising strengths. Zoom in for the best view.}
    \label{fig:visual_enhancement}
\end{figure*}

\begin{figure*}[t]
    \centering
    \includegraphics[width=1\linewidth]{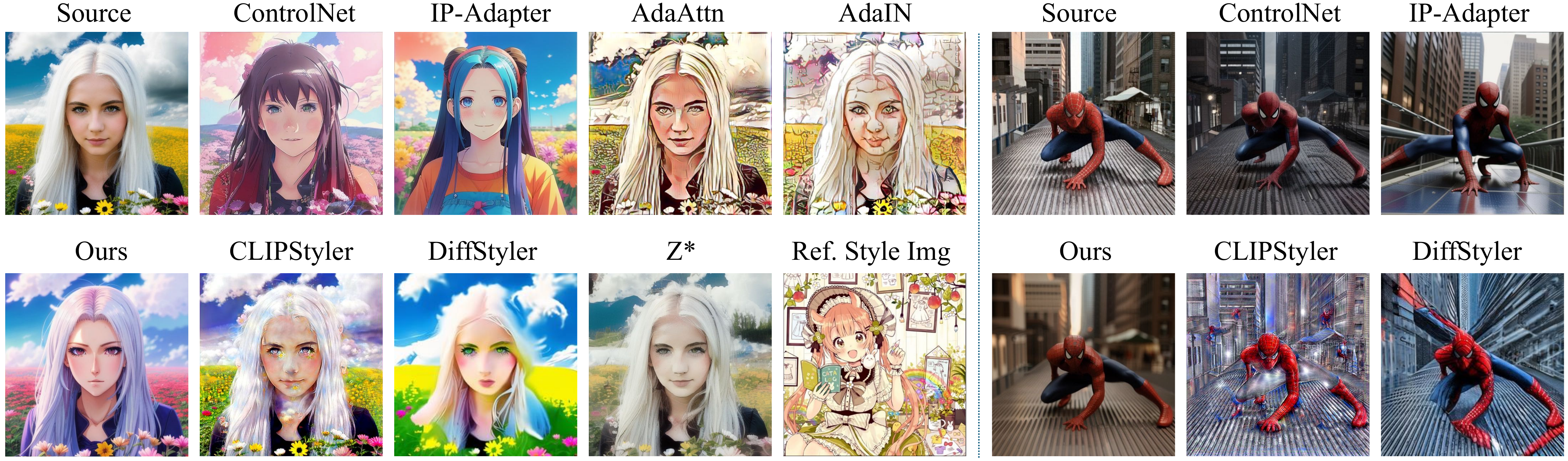}

    \caption{\resp{Qualitative comparisons with various style transfer methods. Left: Anime style transfer, where AdaIN, AdaAttn, and Z* are style-image-driven style transfer methods.
    right: Background blur for visual enhancement,which image-driven methods fail to achieve due to the lack of a reference image.}}
    \label{fig: exp4}
\end{figure*}

\subsection{Implementation Details}

\minor{We conduct comparison experiments by loading the open-source Realistic-Vision-V4 weights~\cite{realistic} into the Stable Diffusion V1 model, which generates images closely resembling those in the real world. }
We use DDIM deterministic sampling with 50 steps in all of our experiments.
\resp{In the case of real guidance images, we conduct deterministic DDIM inversion with $1,000$ forward steps, followed by deterministic DDIM sampling with 1000 backward steps to achieve a relative precise inversion. 
Our stylized results are performed with 50 sampling timesteps, thus we save the corresponding latents every 25 timesteps for use in each step of the 50 timesteps style transfer process.}
In addition, we set the classifier-free guidance scale for real and generated guidance images to 7.5. 
The ASI module only replaces the cross-attention module of the diffusion model in the conditional branch. 
\resp{For Stable Diffusion V1 model, we set the default thresholds for the spatial mask to $\alpha=0.7$, and set the attention head mask to $n=6$ and $n=7$ for visual enhancement and style transfer tasks, respectively.}
\resp{For Stable Diffusion XL model, we set the $\alpha=0.6$, $n=16$ and $n=18$ for visual enhancement and style transfer tasks, respectively.}
\resp{In addition, following the PNP~\cite{tumanyan2023plug}, during the editing process, we replace the self-attention scores from layers 4 to 11 of the UNet with the corresponding self-attention scores from the reference diffusion process for the first 25 timesteps.
Similarly, we also replace the convolution features from layer 4 of the UNet during the first 40 timesteps.}

\subsection{Qualitative Comparison}

\subsubsection{Style Transfer}
\label{exp1}

To verify the effectiveness and generalization of ASI, we show various style transfer results applied by our approach for real-world images and generated images.
%
For each result, we show images before and after stylized.
As shown in Fig.~\ref{fig:style_transfer}, we compare our method with SOTA text-guided style transfer methods based on diffusion models, including FreeDiff~\cite{wu2024freediff}, P2P~\cite{hertz2022prompt}, PNP~\cite{tumanyan2023plug}, and SDEdit~\cite{meng2021sdedit}.
\resp{The first three rows and the last three rows respectively show the results of style transfer based on real and generated images.}
Since our method aims to incorporate style information at the fine-grained feature level, compared to prompt-level style injection, it allows us to stylize in the smaller local region of images, thereby reducing the distortion of image structure.
\resp{Although SDEdit is simple but effective, it requires manual adjustment of the trade-off between the two.
%
FreeDiff can preserve object consistency in stylized images, but it struggles to accurately maintain the image's structural consistency (Row.~1,2,4-6) and effective stylization (Row.~3).
PNP and P2P either have excessive stylization(e.g., Row.~1,2,4, and 5) leading to structural and instance deviation, or they lack effective stylization(Row.~3 and 4).
As shown in Col.~2 of Fig.~\ref{fig:style_transfer}, our method effectively stylizes images while maintaining structural consistency with the source content image, achieving a good balance between stylization and structural preservation compared to other SOTA methods.}

\subsubsection{Visual Enhancement}
\label{exp2}
Benefiting from the text-guided style transfer method instead of the image-guided one, the diffusion-based method can be expanded to application areas unreachable by traditional image-to-image style transfer methods~\cite{zhu2017unpaired, huang2017arbitrary, gatys2015neural}, particularly those related to visual enhancements such as realism, depth of field, and other photographic techniques.
This is because these "styles" cannot be represented through an image.

As shown in Fig.~\ref{fig:visual_enhancement}, while FreeDiff~\cite{wu2024freediff}, P2P~\cite{hertz2022prompt}, PNP~\cite{tumanyan2023plug}, and SDEdit~\cite{meng2021sdedit} are based on the diffusion model, they only preserve the overall structure of the source image and cannot achieve effective visual enhancement according to the given prompt.
This limits the application of the above methods to this specific task as it requires a more strict consistency between the source and target images.

Thanks to our proposed fine-grained style incorporation idea, our method efficiently enhances the visual quality of images while maintaining structural consistency. 
\resp{The first three rows and the last two rows respectively of the Col.~2 of Fig.~\ref{fig:visual_enhancement} show the results of style transfer based on real and generated images.
In terms of appearance enhancement, our method better maintains the object's semantic and structure (Row.~1 and 2).
In addition, our method can perform certain photographic-level image adjustments, such as background blur and long-exposure effects~(Row.~3 and 4), and can also adjust the image's tones~(Row.~5). }
%
These impressive results reveal the immense potential of diffusion models in addressing low-level vision tasks.

\begin{table*}[t]
\centering
\caption{ Quantitative evaluation for stylized images across style transfer and visual enhancement tasks. We report structure distance and human preference scores for our method and other baselines. }
\begin{tabular}{cc|ccccc|ccccc} 
\toprule
 &Metric& \multicolumn{5}{c|}{Structure Distance~$\downarrow$}& \multicolumn{5}{c}{Human Preference~$\uparrow$} \\
\midrule
 &Method&  \bf    Ours & FreeDiff & P2P & PNP & SDEdit &  \bf    Ours & FreeDiff & P2P & PNP & SDEdit \\
\midrule
&Style Transfer&    \bf  \resp{0.016} & 0.071 & 0.053& 0.054 & 0.087&   \bf \resp{70\%} & \resp{4\%} & 6\% & 16\% & 4\%
  \\
&Visual Enhancement&   \bf  0.011 & 0.066& 0.049& 0.041 & 0.072&  \bf 78\% & 6\% &  8\% & 8\% & 0\%      \\

\bottomrule
\end{tabular}
\label{tab:evaluation1}
\end{table*}

\begin{table}[!t] 
\centering
\caption{ CLIP Aesthetic Score Comparisons on Visual Enhancement.}

\begin{tabular}{c|cccccc}
\toprule
&  Source &  \bf Ours &  FreeDiff &  P2P &  PNP &  SDEdit \\
\midrule
CLIP Aes. $\uparrow$ &  5.46 &  \bf 6.13 &  5.37 &  4.90 &   5.57 &  5.25 \\

\bottomrule
\end{tabular}
\label{tab:evaluation2}
\end{table}

\begin{table}[!t]
    \centering
    \caption{CLIP Score Comparison on Style Transfer.}
    \begin{tabular}{c|ccccc}
    \toprule
      &  \bf Ours & FreeDiff & P2P & PNP & SDEdit\\
    \midrule
    CLIP Score $\uparrow$ & \bf 0.352 & 0.346 & 0.348 & 0.330 & 0.329  \\ 
    BG LPIPS $\downarrow$ & \bf {0.157} & 0.160 &  0.163 & 0.158 & 0.177 \\ 
    \bottomrule
    \end{tabular}
    \label{tab:clip}
\end{table}

\subsubsection{Compare with More Style Transfer Methods}

We further qualitatively compare our method with other methods that have different experimental settings, including the diffusion-based fine-tuning methods ControlNet~\cite{zhang2023adding}, IP-Adapter~\cite{ye2023ip}, DiffStyler~\cite{huang2024diffstyler}, the text-driven style transfer method based on the CLIP model~\cite{radford2021learning} \minor{ClipStyler}~\cite{kwon2022clipstyler}, and the traditional image-driven style transfer methods such as \minor{AdaIN}~\cite{huang2017arbitrary}, AdaAttn~\cite{liu2021adaattn} and Z*~\cite{deng2024z}.
As shown in Fig.~\ref{fig: exp4}, compared to the aforementioned methods, our approach yields more vibrant results in the anime-style transformation task.

Furthermore, for visual enhancement, traditional image-driven style transfer methods \minor{AdaIN}~\cite{huang2017arbitrary}, AdaAttn~\cite{liu2021adaattn} and Z* fall short as it cannot achieve this effect due to the unavailability of the corresponding reference style image. Simultaneously, our method surpasses text-driven style transfer methods ControlNet~\cite{zhang2023adding}, IP-Adapter~\cite{ye2023ip}, DiffStyler~\cite{huang2024diffstyler}, and \minor{ClipStyler}~\cite{kwon2022clipstyler}, as it better enhances image's expressiveness while preserving the structure of the image.

\subsection{Quantitative Comparison}

\resp{To better evaluate our method, we use multiple quantitative metrics in ImageNet-R-TI2I~\cite{tumanyan2023plug} dataset for assessment, the results of which are presented in Tab.~\ref{tab:evaluation1}. 
We evaluate the average performance of seven common style transfers, such as ``Anime artwork'', ``Watercolor painting'', ``Play-doh'', ``Origami'', ``Pixel-art'', and ``Cyberpunk'' in real-world images for the style transfer task. For visual enhancement in real-world images, we employ five style prompts such as ``Cinematic'', ``Realistic'' and ``Best quality''.}
%

As shown in Tab.~\ref{tab:evaluation1}, we utilize Structure Distance~\cite{tumanyan2022splicing, parmar2023zero} to measure the structural consistency between the stylized image and the source content image. This measurement is computed by calculating the cosine distance between the two using DINO embeddings~\cite{caron2021emerging}.
\minor{A lower score on Structure Distance means that the structure of the stylized image is more similar to the source content image.}

\minor{We also present the human preference score in Tab.~\ref{tab:evaluation1}}. This user study involved a survey of 40 participants, all of whom hold bachelor's degrees or higher. They were asked to select the best method based on structural consistency and style transfer effects. Each participant was required to evaluate the results of five randomly assigned experimental groups. From the 200 votes collected, it is evident that our method holds significant advantages over other methods.

In addition, we present CLIP Aesthetic Score~\cite{schuhmann2022laion} of visual enhancement on real-world images in Tab.~\ref{tab:evaluation2}.
A higher CLIP Aesthetic Score indicates that the fusion of style and content is more aesthetically pleasing. 
We also present the CLIP score in Tab.~\ref{tab:clip} to measure the correspondence between style prompts and stylized images. 
\minor{ Lastly, to ensure that the background is well preserved after stylization, we report the background LPIPS distance (BG LPIPS) in Tab.~\ref{tab:clip} for successfully stylized images. This metric is computed by measuring the LPIPS distance between the background regions of the source and stylized images, identified using a detector. A lower BG LPIPS score indicates better preservation of the source image's background. }
The results show the superiority of our method in following textual requirements through stylized image outputs.

\begin{figure}[t]
    \centering
    \includegraphics[width=1\linewidth]{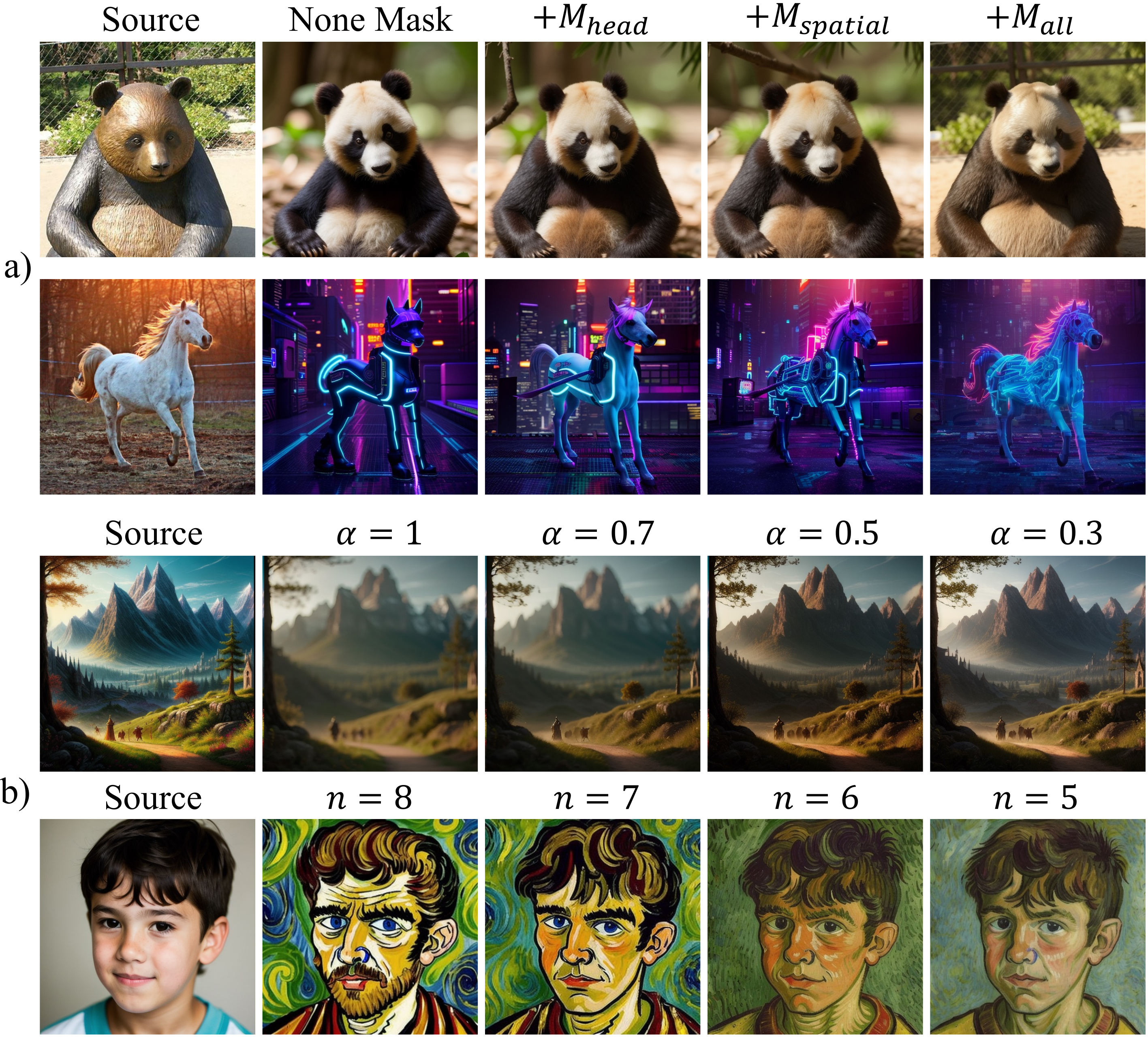}
    \caption{Ablation study on the guidance mask of the AdaBlending module.}
    \label{fig: abl}
\end{figure}

\subsection{Ablation Study}

\resp{
We conduct ablation experiments on the guidance mask of the AdaBlending module. 
As shown in Fig.~\ref{fig: abl} (a)'s anime and neonpunk style trasfer task, noticeable structural drifts are observed when no guidance mask is used.
When only $M_{head}$ is used, the overall structure of the image is preserved, but drifts occur in the local details of the image.
However, when both spatial and attention-head level masks are used concurrently, AdaBlending delivers consistent structural and semantic style transfer results. These results highlight the crucial role of guidance masks in achieving accurate and consistent style transfer outcomes.}

\minor{
Furthermore, we conducted an ablation study on the hyperparameters $\alpha$ and $n$, varying one while keeping the other constant. }
As shown in Row.~1 of Fig.~\ref{fig: abl} (b) for the cinematic transfer task, setting $\alpha=0.7$ achieves effective blurring and tone adjustment while preserving clear semantic content in the image. 
The Row.~2 of Fig. \ref{fig: abl} presents a van Gogh painting transfer task. It can be seen that the stylized image with $n=8$ (without the attention-head mask) exhibits semantic drift compared to the source image. In contrast, when $n=6$, the image has suitable stylistic textural details but lacks color contrast to emphasize the distinct style. However, when $n=7$, the stylized image retains the source image's structure and achieves effective style transfer.

\section{Conclusion}
In this work, we aim to study text-driven style transfer.
To mitigate the well-known issue of structure inconsistency induced by text-image misalignment in diffusion-based T2I models, we propose a novel and effective tuning-free algorithm, termed Adaptive Style Incorporation (ASI), to incorporate the style information on the fine-grained feature level. 
It consists of the Siamese Cross-Attention (SiCA) to decouple the single-track cross-attention to a dual-track structure to obtain separate content and style features and the Adaptive Content-Style Blending (AdaBlending) module to couple the content and style information from a structure-consistent manner.
We conduct a thorough evaluation of our method across diverse source image domains, both real and generated ones, on a range of common style transfer tasks.
%
%
Experimental results have demonstrated the effectiveness of our proposed ASI and exhibit significant advantages over SOTA methods.


\bibliographystyle{IEEEtran}
\bibliography{reference}

\clearpage

\begin{IEEEbiography}
[{\includegraphics[width=1in,height=1.25in,clip,keepaspectratio]{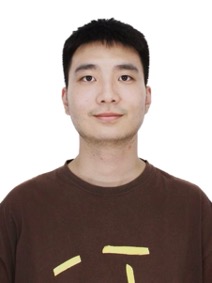}}]{Yanqi Ge} received his M.S. degree from the School of Computer Science and Engineering, University of Electronic Science and Technology of China, in 2024. His research interests include transfer learning, semantic segmentation and image generation.
\end{IEEEbiography}
\begin{IEEEbiography}
[{\includegraphics[width=1in,height=1.25in,clip,keepaspectratio]{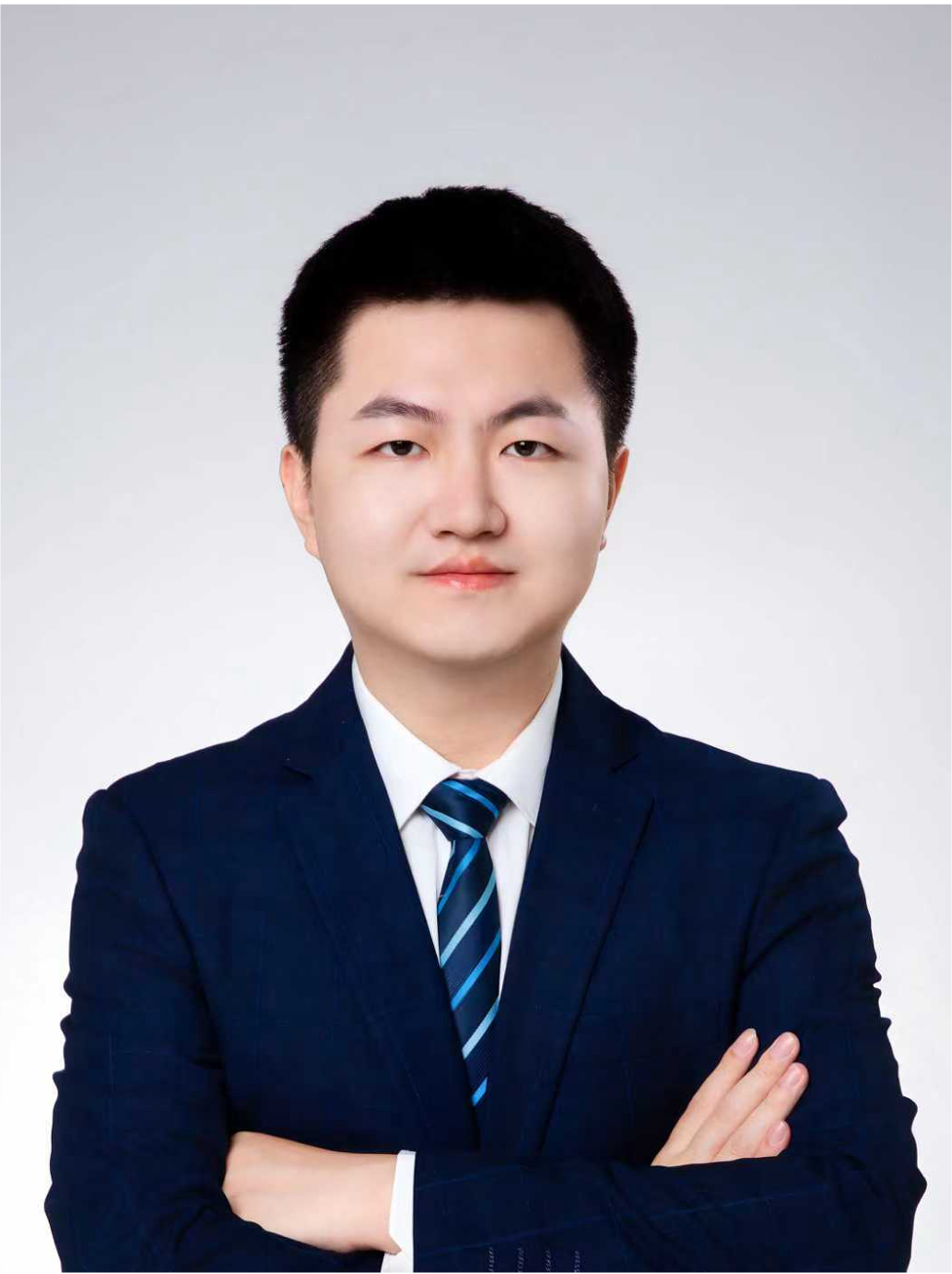}}]{Jiaqi Liu} received his B.S. degree from Dalian University of Technology in 2019 and his M.S. degree from the Southern University of Science and Technology, China, in 2024. His research interests include anomaly detection and image editing.
\end{IEEEbiography}
\begin{IEEEbiography}
[{\includegraphics[width=1in,height=1.25in,clip,keepaspectratio]{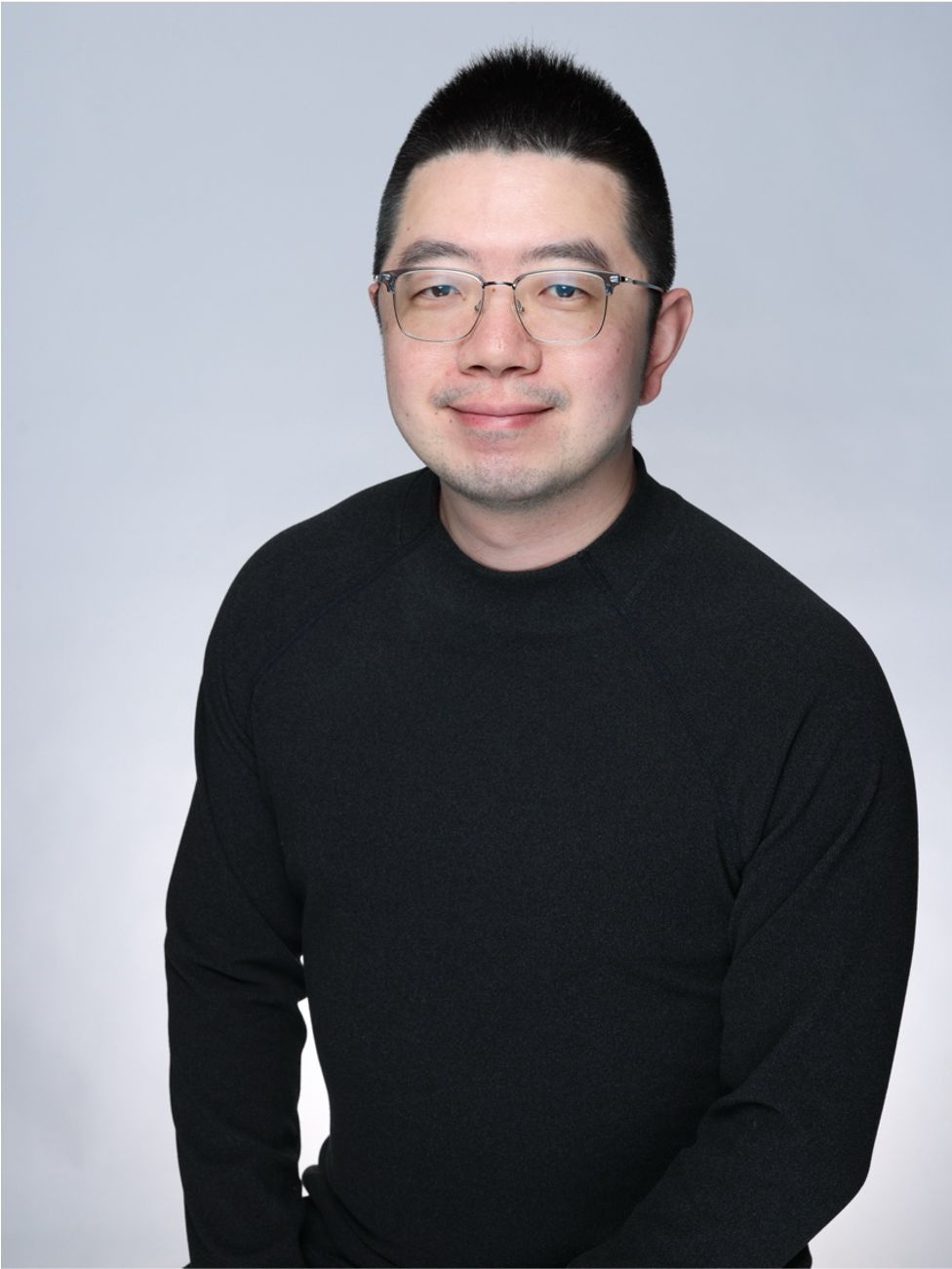}}]{Qingnan Fan} is a Lead Researcher and team manager at the Imaging Algorithm Center, VIVO, where he drives core algorithms for flagship smartphone photography using 3D and AIGC technologies. He was a Senior Researcher at Tencent AI Lab and a postdoc at Stanford with Prof. Leonidas Guibas. He received his PhD in Computer Science from Shandong University under Prof. Baoquan Chen in 2019. His research spans computational photography, computer graphics, 3D vision, and embodied AI.
\end{IEEEbiography}
\begin{IEEEbiography}
[{\includegraphics[width=1in,height=1.25in,clip,keepaspectratio]{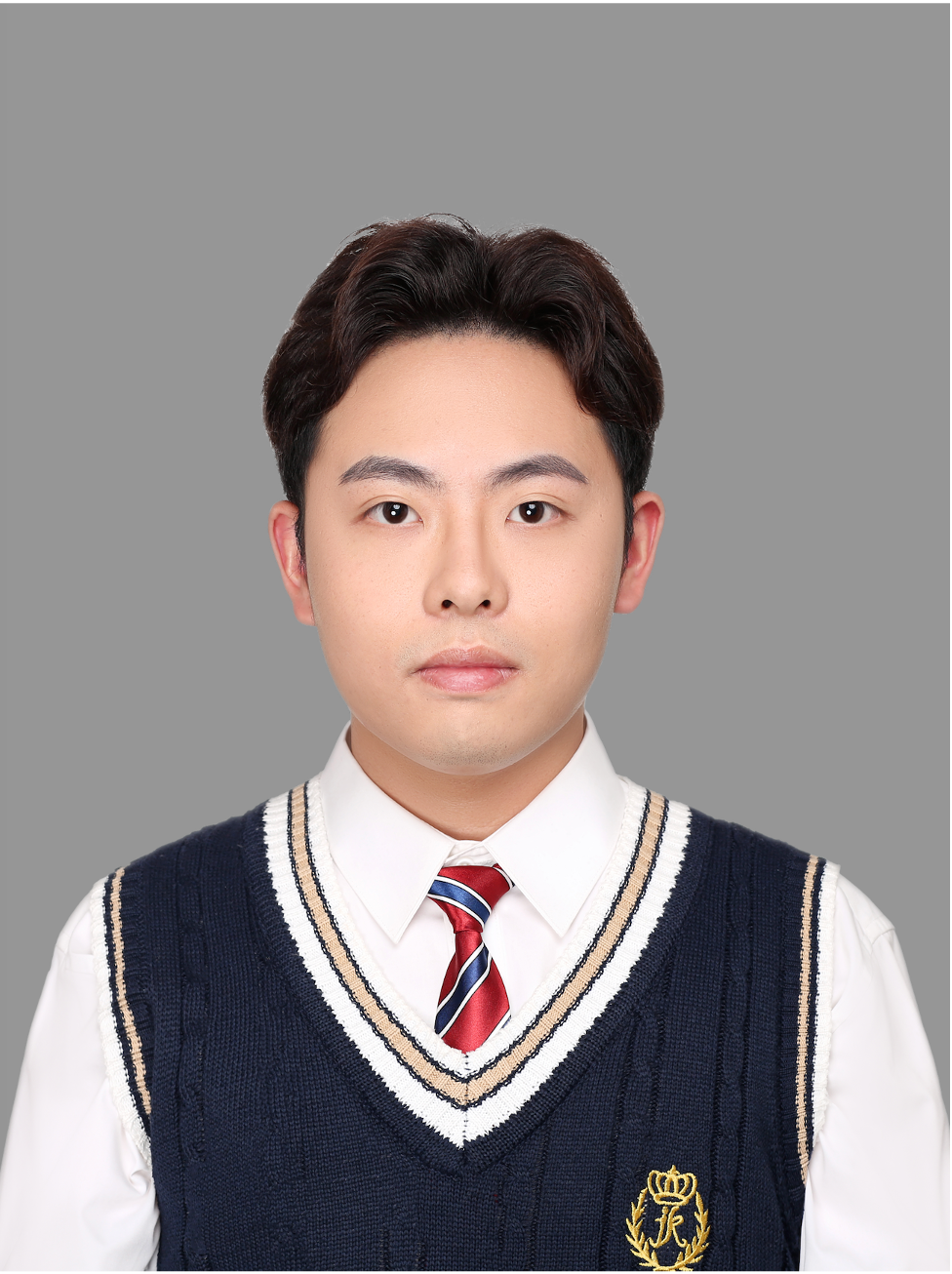}}]{Xi Jiang} received his B.S. degree from Xi’an Jiaotong University, China, in 2020 and his M.S. degree from the Southern University of Science and Technology, China, in 2023. He is currently a Ph.D. candidate in the Department of Computer Science and Engineering at Southern University of Science and Technology. His research interests include computer vision and machine learning.
\end{IEEEbiography}
\begin{IEEEbiography}
[{\includegraphics[width=1in,height=1.25in,clip,keepaspectratio]{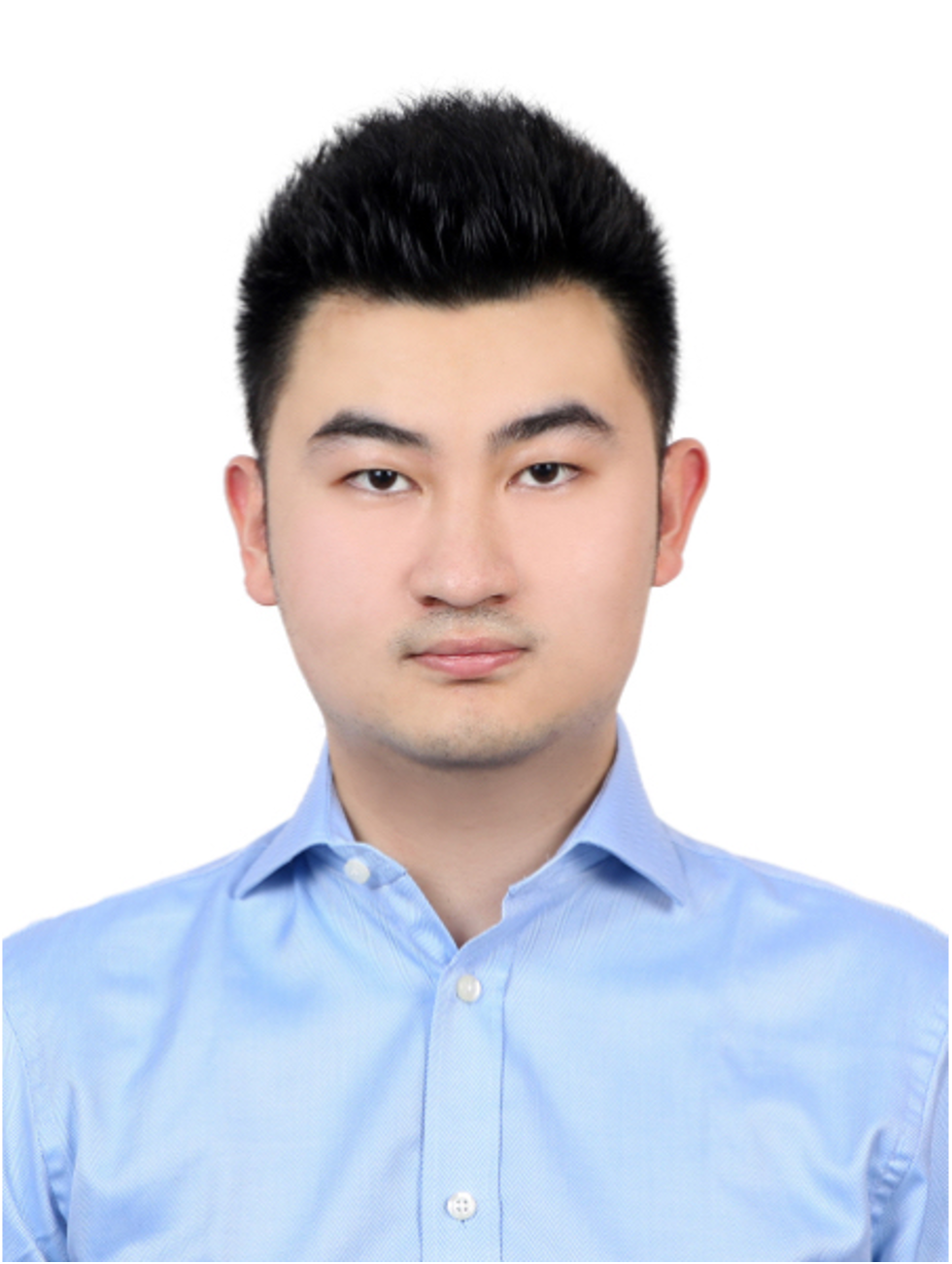}}]{Ye Huang} is Associate Professor at the University of Electronic Science and Technology of China, where he completed his post-doctoral training. His research focuses on AGI representation learning and AI4Science; earlier work on semantic segmentation holds top accuracy on major benchmarks. He earned his Ph.D. and B.E. in computer science from the University of Technology Sydney.
\end{IEEEbiography}
\begin{IEEEbiography}
[{\includegraphics[width=1in,height=1.25in,clip,keepaspectratio]{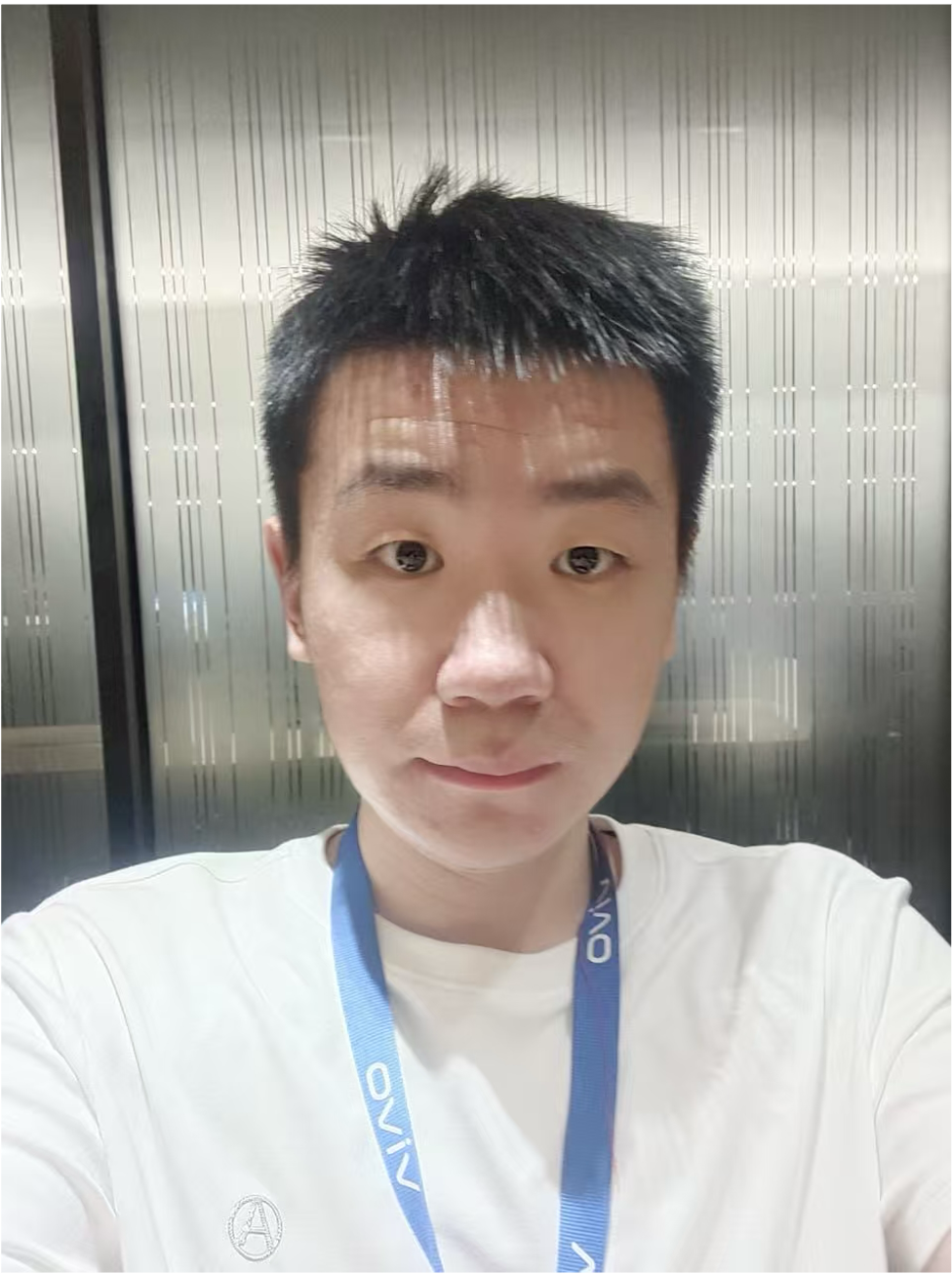}}]{Shuai Qin} received his M.S. degree from the National Key Laboratory of Wireless Communications from the University of Electronic Science and Technology of China in Chengdu.
\end{IEEEbiography}
\begin{IEEEbiography}
[{\includegraphics[width=1in,height=1.25in,clip,keepaspectratio]{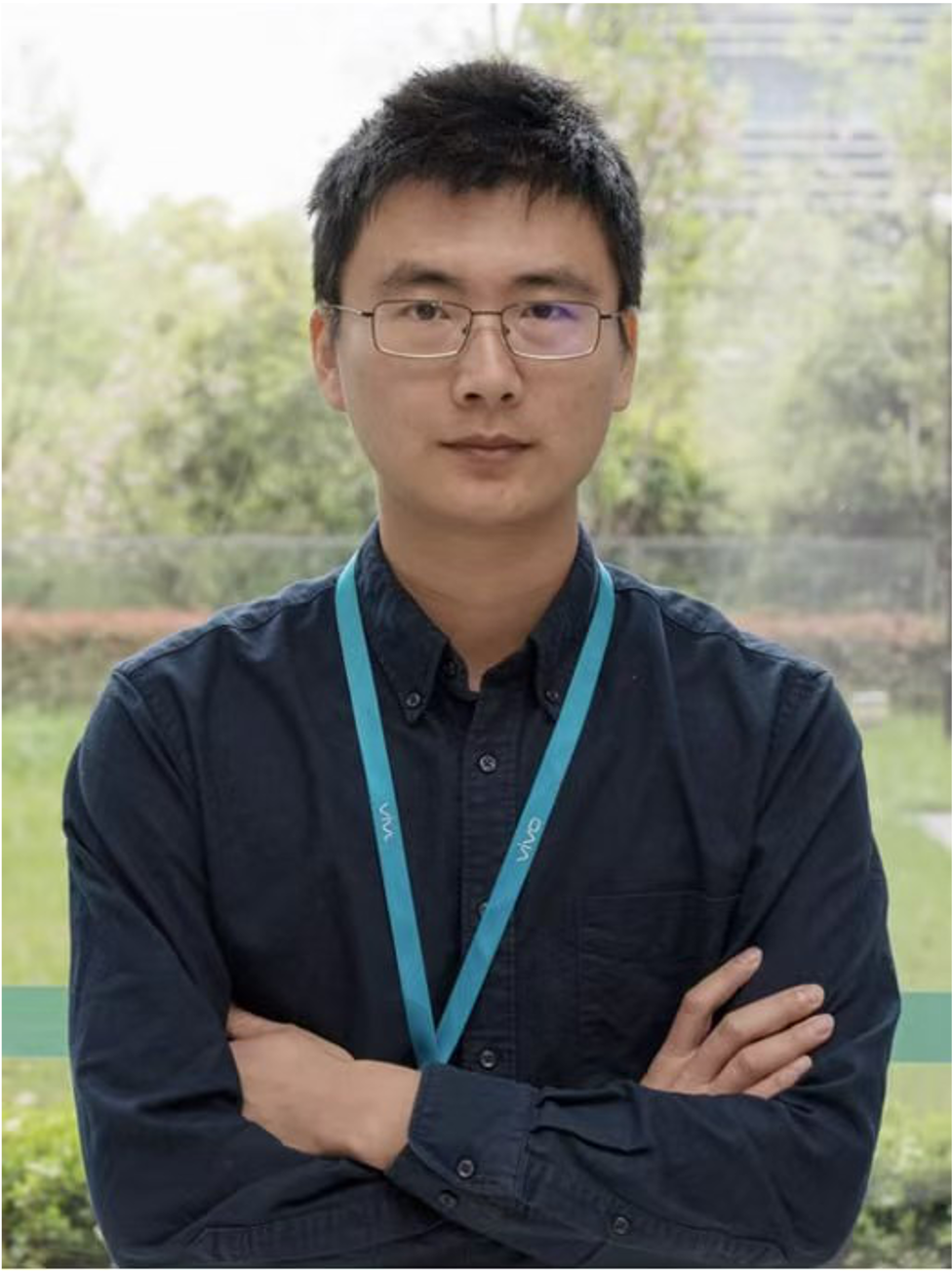}}]{Hong Gu} received the B.Eng. and Ph.D. degrees in automation from Zhejiang University, Hangzhou, China, in 2006 and 2011, respectively. He is currently with VIVO Mobile Communication Company Ltd., Hangzhou. His research interests are computational photography, computer vision, and deep learning.
\end{IEEEbiography}
\begin{IEEEbiography}
[{\includegraphics[width=1in,height=1.25in,clip,keepaspectratio]{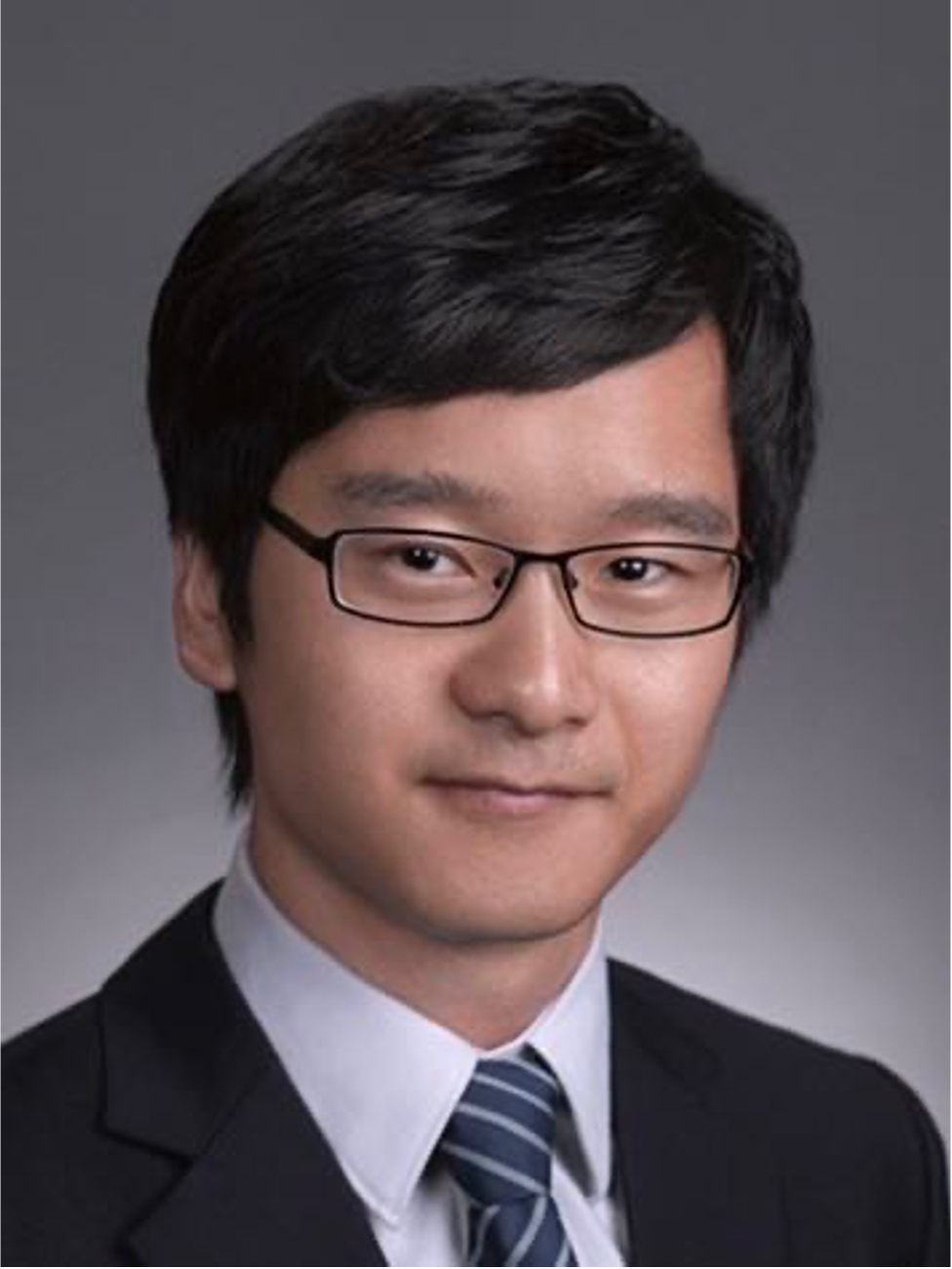}}]{Wen Li}  received the B.E. and M.S. degrees from the Beijing Normal University in 2007 and 2010, respectively, and the Ph.D. degree from the Nanyang Technological University in 2015. He currently is a Full Professor at the Shenzhen Institute for Advanced Study and the School of Computer Science and Engineering, University of Electronic Science and Technology of China. His main research interests include transfer learning, multi-view learning, multiple kernel learning, and their applications in computer vision.
\end{IEEEbiography}
\begin{IEEEbiography}
[{\includegraphics[width=1in,height=1.25in,clip,keepaspectratio]{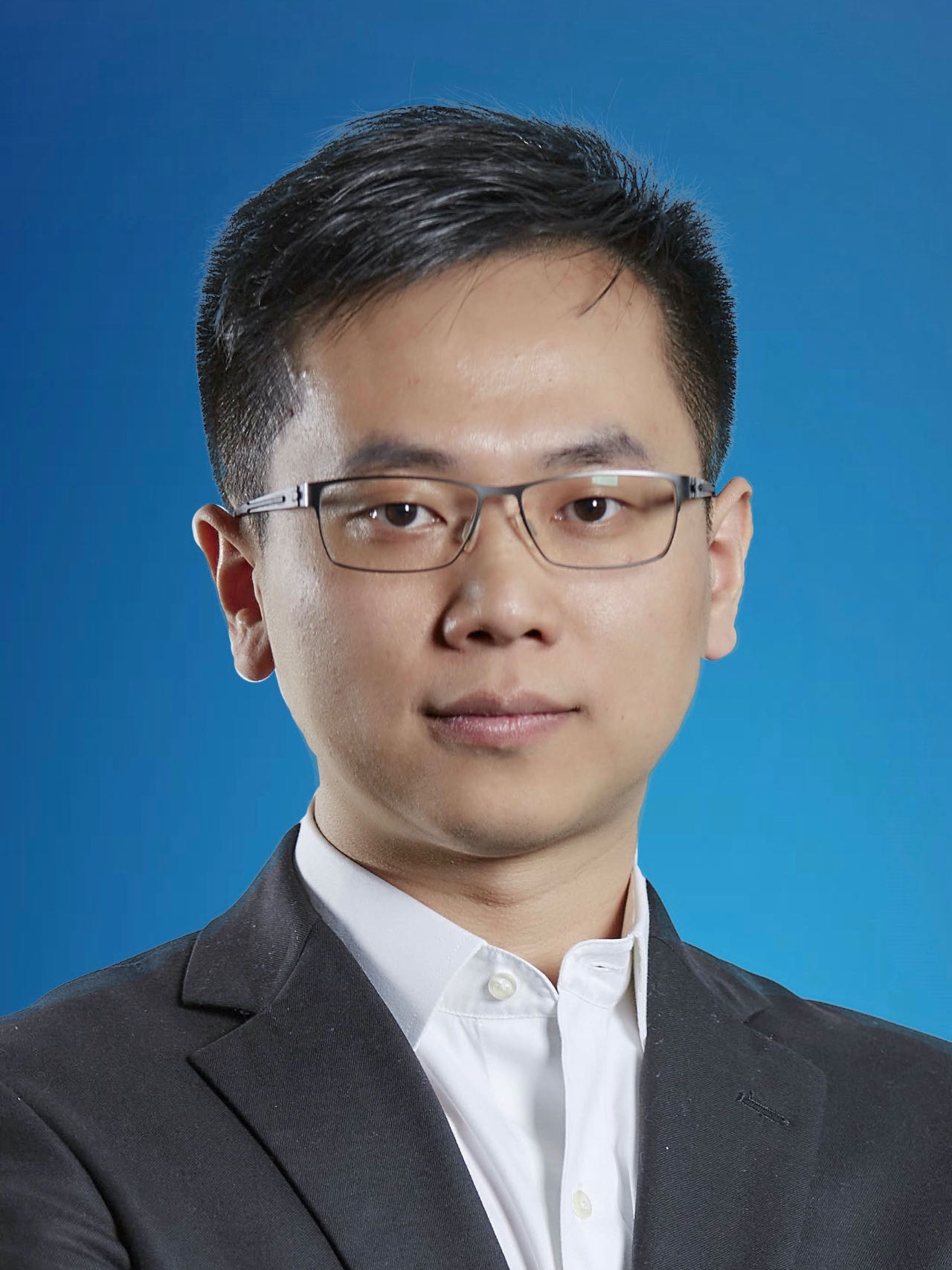}}]{Lixin Duan} received the B.E. degree from the University of Science and Technology of China in 2008, and the Ph.D. degree from the Nanyang Technological University in 2012. He currently is a Full Professor at the School of Computer Science and Engineering, University of Electronic Science and Technology of China. His main research interests include machine learning algorithms (especially in transfer learning and domain adaptation) and their applications in computer vision. He was a recipient of the Microsoft Research Asia Fellowship in 2009 and the Best Student Paper Award at the IEEE Conference on Computer Vision and Pattern Recognition 2010.
\end{IEEEbiography}

\clearpage

\appendix

\section*{More Style Transfer Experiments}
\begin{figure*}[t]
    \centering
    \includegraphics[width=1\linewidth]{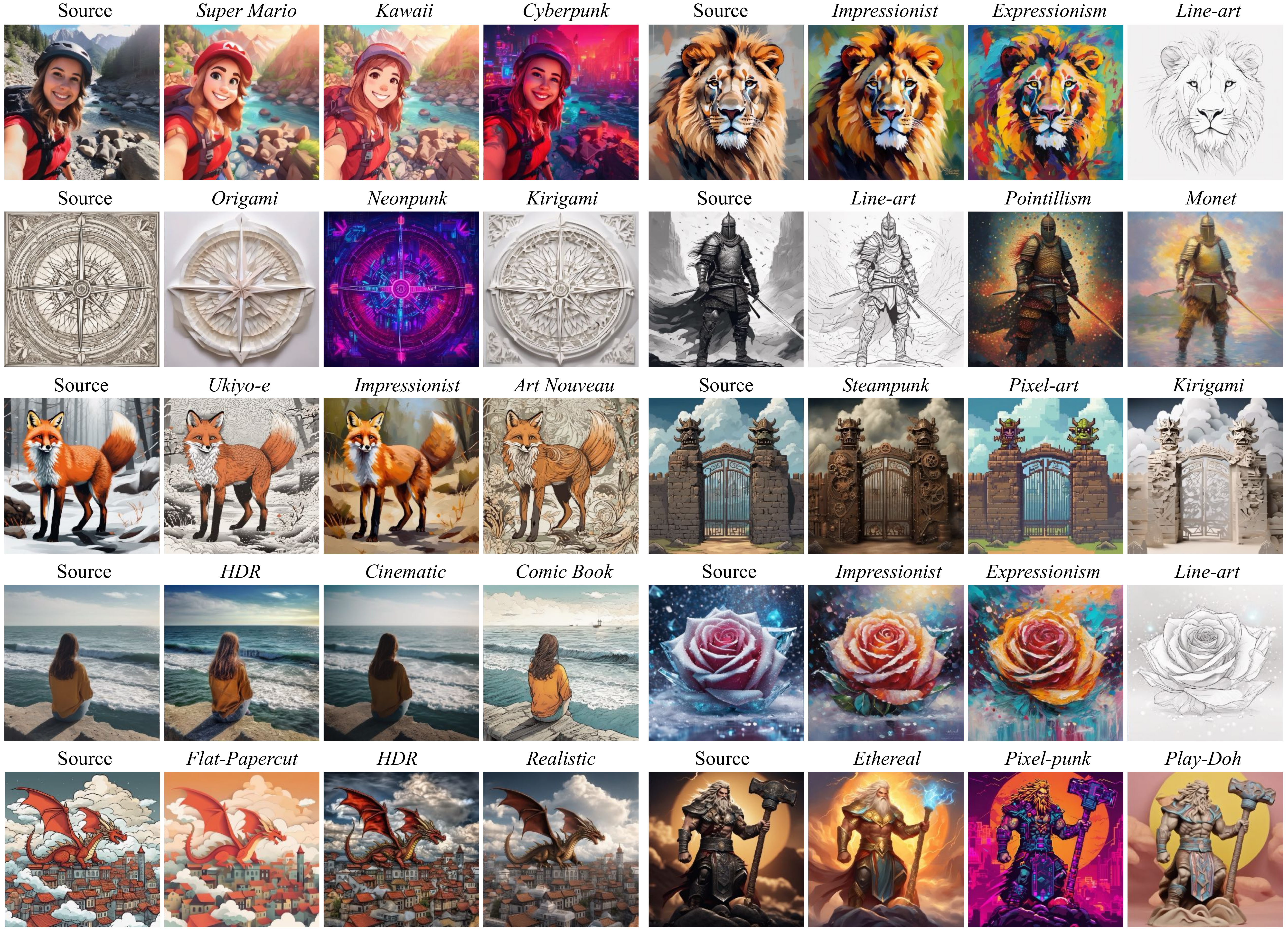}
    \caption{More stylization examples of our methods.}
    \label{fig:1}
\end{figure*}

We present more of our experimental results based on SDXL in Fig.~\ref{fig:1}.

\section*{More Ablation Study}
\begin{figure*}[t]
    \centering
      \includegraphics[width=0.75\linewidth]{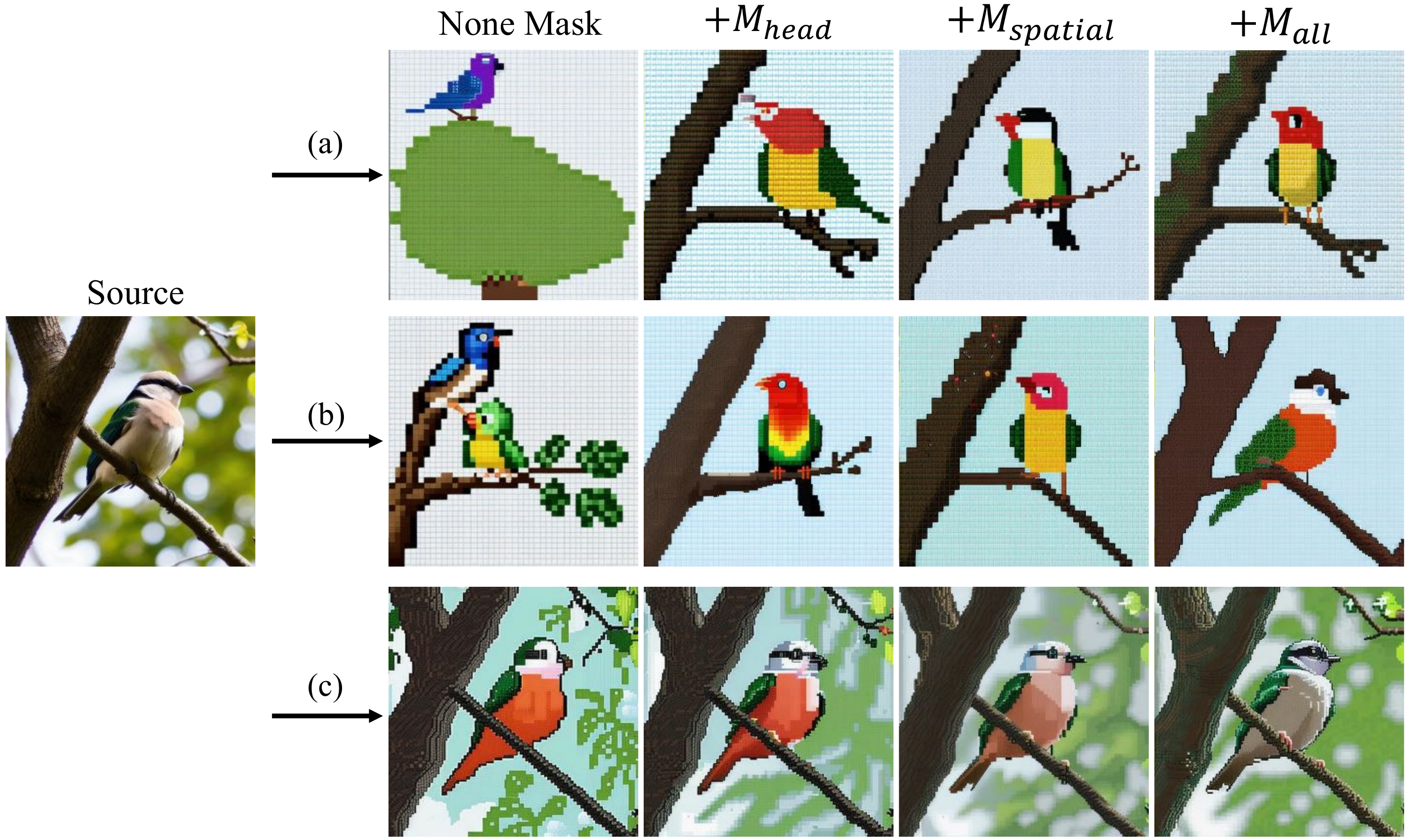}
    \caption{Ablation study of the proposed method under three reference-diffusion feature-reuse schemes: (a) no feature reuse; (b) reuse of features from only the first five timesteps; (c) the standard scheme. }
    \label{figure: s6}
\end{figure*}

\minor{To further explore the effect of guidance masks of the proposed method, we conduct ablation experiments under the three reference-diffusion reuse schemes shown in Fig.~\ref{figure: s6}: (a) no feature reuse, (b) reuse of features from only the first five timesteps, and (c) the standard scheme. Across all settings, introducing $M_{head}$ or $M_{spatial}$ markedly improves structural consistency relative to the source image. When both guidance masks are used together, the proposed method achieves robust structural alignment and faithful semantic stylization.}

\section*{Limitations}

\begin{figure}[t]
    \centering
    \includegraphics[width=1\linewidth]{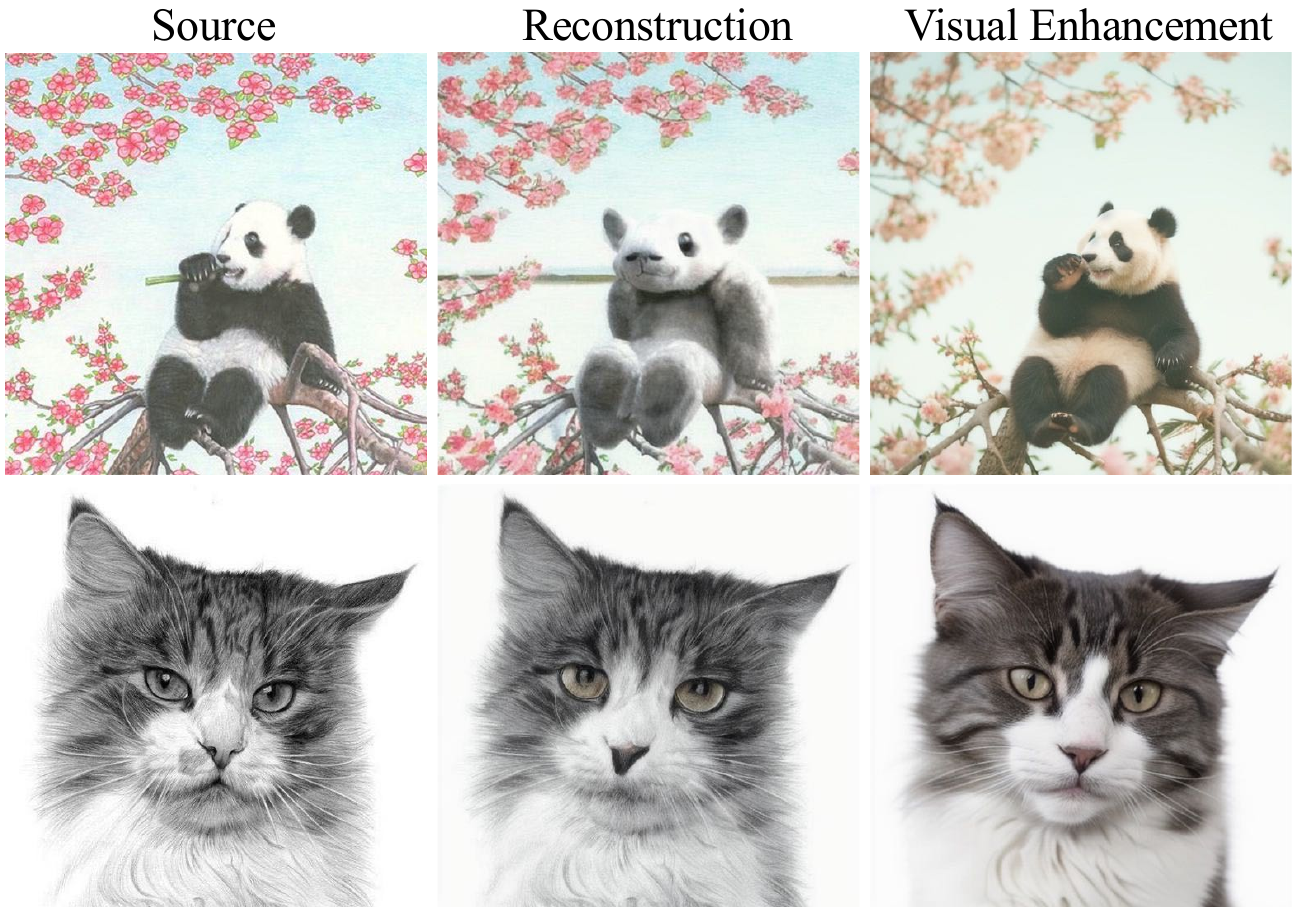}
    \caption{Typical defective cases caused by reconstruction failure.}
    \label{fig:limit1}
\end{figure}

\begin{table}[t]
    \centering
    \caption{Computation Cost Comparison on Stylized Process.}
    \begin{tabular}{c|ccc}
    \toprule
      &  \bf Ours & P2P & PNP \\
    \midrule
    Total Time & 16.5 & 13.2 & 14.1    \\ 
    CUDA Time & 12.7 & 9.3 & 10.1 \\
    VRAM & 7326M & 5276M & 7324M \\ 
    \bottomrule
    \end{tabular}
    \label{tab:cost}
\end{table}

Although our proposed approach achieves impressive performance on style transfer and visual enhancement tasks, it still has the following three limitations: Firstly, due to the inability of the DDIM inversion strategy to perfectly reconstruct the source image, our stylized results may lose some details of the source image, such as the bamboo in the panda's hand and the cat's eyes in Fig.~\ref{fig:limit1}. 
Moreover, since our method is designed to achieve structural consistency, it becomes challenging to achieve unique styles expressed through structural distortion, such as abstract styles and Picasso-like Cubist styles.
In addition, we calculate the stylized process computation cost of our method and P2P, PNP based on an A10 GPU, which are presented in Tab.~\ref{tab:cost}. Since our method requires the computation of feature covariance and principal components during the stylization process, it slightly increases the computation cost compared to P2P and PNP.

\section*{Text Prompt Used in Quantitative Comparison}

Contrasting with prior research that typically employs simple style phrases for style transfer, we utilize style prompts officially supplied by Stable Diffusion~\cite{stablediffusionstyle} to ensure more formal evaluations.
The positive and negative prompts used during the style transfer process are described as follows:

\textbf{Anime artwork}. Positive: \textit{anime artwork \{prompt\}. anime style, key visual, vibrant, studio anime, highly detailed}.
Negative: \textit{photo, deformed, black and white, realism, disfigured, low contrast. }

\textbf{Watercolor painting} Positive: \textit{watercolor painting\{prompt\}. vibrant, beautiful, painterly, detailed, textural, artistic}. 
Negative: \textit{anime, photorealistic, 35mm film, deformed, glitch, low contrast, noisy.}

\textbf{Play-doh style} Positive: \textit{style prompt: play-doh style \{prompt\}. sculpture, clay art, centered composition, Claymation.} 
Negative:  \textit{sloppy, messy, grainy, highly detailed, ultra textured, photo.}

\textbf{Origami} Positive: \textit{ origami style \{prompt\}. paper art, pleated paper, folded, origami art, pleats, cut and fold, centered composition.} 
Negative:  \textit{noisy, sloppy, messy, grainy, highly detailed, ultra textured, photo.}

\textbf{Pixel-art style}: Positive: \textit{pixel-art \{prompt\}. low-res, blocky, pixel art style, 8-bit graphics. }
Negative: \textit{sloppy, messy, blurry, noisy, highly detailed, ultra textured, photo, realistic.}

\textbf{Cyberpunk} Positive: \textit{cyberpunk cityscape \{prompt\}. neon lights, dark alleys, skyscrapers, futuristic, vibrant colors, high contrast, highly detailed}. 
Negative:  \textit{natural, rural, deformed, low contrast, black and white, sketch, watercolor.}

\textbf{Cinematic}. Positive: \textit{cinematic film still \{prompt\}. shallow depth of field, vignette, highly detailed, high budget, bokeh, cinemascope, moody, epic, gorgeous, film grain, grainy.} 
Negative:  \textit{anime, cartoon, graphic, text, painting, crayon, graphite, abstract, glitch, deformed, mutated, ugly, disfigured.}

\textbf{Best-quality} Positive: \textit{ breathtaking \{prompt\} . award-winning, professional, highly detailed}
Negative:  \textit{ugly, deformed, noisy, blurry, distorted, grainy}

\textbf{Realistic} Positive: \textit{ hyperrealistic art \{prompt\}. extremely high-resolution details, photographic, realism pushed to extreme, fine texture, incredibly lifelike}
Negative:  \textit{ugly, deformed, noisy, blurry, distorted, grainy}

\end{document}